\documentclass[twoside]{article}

\usepackage[accepted]{aistats2019}

\usepackage{natbib}
\setcitestyle{authoryear}
\usepackage[utf8]{inputenc} 
\usepackage[T1]{fontenc}    
\usepackage[draft]{hyperref}       
\usepackage{url}            
\usepackage{booktabs}       
\usepackage{amsfonts}       
\usepackage{nicefrac}       
\usepackage{microtype}      
\usepackage{wrapfig}
\usepackage{placeins}

\usepackage{algorithm}
\usepackage{algorithmic}

\usepackage{xcolor}
\usepackage{amsmath}
\usepackage{amssymb}
\usepackage{chngcntr}
\usepackage{amsthm}
\usepackage{ifthen}
\usepackage{graphicx}
\usepackage{subfigure}

\makeatletter
\newtheorem*{rep@theorem}{\rep@title}
\newcommand{\newreptheorem}[2]{%
\newenvironment{rep#1}[1]{%
 \def\rep@title{#2 \ref{##1}}%
 \begin{rep@theorem}}%
 {\end{rep@theorem}}}
\makeatother

\newtheorem{lemma}{Lemma}
\newtheorem{proposition}{Proposition}
\newtheorem{defn}{Definition}

\newtheorem{corollary}{Corollary}
\newtheorem{assumption}{Assumption}
\newreptheorem{lemma}{Lemma}
\newreptheorem{assumption}{Assumption}
\newreptheorem{proposition}{Proposition}
\newreptheorem{defn}{Definition}
\newreptheorem{conjecture}{Conjecture}
\newreptheorem{corollary}{Corollary}

\newcommand{\hide}[1]{}

\newcommand\sN{\ensuremath{\mathcal{N}}}

\newcommand\bi{\ensuremath{\mathbf{i}}}

\newcommand\FigStar[4]{\begin{figure*}[ht] \begin{center} \includegraphics[scale=#2]{#1} \end{center} \caption{\label{fig:#3} #4} \end{figure*}}

\newcommand{\var}{\text{Var}} 
\newcommand{\cov}{\text{Cov}} 

\newcommand\R{\ensuremath{\mathbb{R}}} 
\newcommand\eqdef{\ensuremath{\stackrel{\rm def}{=}}} 
\newcommand{\bone}{\mathbf{1}} 
\newcommand\refeqn[1]{(\ref{eqn:#1})}

\newcommand\refsec[1]{Section~\ref{sec:#1}}

\newcommand\reffig[1]{Figure~\ref{fig:#1}}

\newcommand\reflem[1]{Lemma~\ref{lem:#1}}

\newcommand\refprop[1]{Proposition~\ref{prop:#1}}

\ifthenelse{\isundefined{\definition}}{\newtheorem{definition}{Definition}}{}
\ifthenelse{\isundefined{\assumption}}{\newtheorem{assumption}{Assumption}}{}
\ifthenelse{\isundefined{\hypothesis}}{\newtheorem{hypothesis}{Hypothesis}}{}
\ifthenelse{\isundefined{\proposition}}{\newtheorem{proposition}{Proposition}}{}
\ifthenelse{\isundefined{\theorem}}{\newtheorem{theorem}{Theorem}}{}
\ifthenelse{\isundefined{\lemma}}{\newtheorem{lemma}{Lemma}}{}
\ifthenelse{\isundefined{\corollary}}{\newtheorem{corollary}{Corollary}}{}
\ifthenelse{\isundefined{\alg}}{\newtheorem{alg}{Algorithm}}{}
\ifthenelse{\isundefined{\example}}{\newtheorem{example}{Example}}{}
\newcommand{\E}{\ensuremath{\mathbb{E}}} 

\newcommand{\diffmap}{q}
\newcommand{\lineartransformation}{a}
\newcommand{\corrbetweenmodels}{\rho_r}

\newcommand{\llon}{\lambda_{\text{lon}}}
\renewcommand{\i}{{(i)}}
\newcommand{\xti}{x_t^\i}
\newcommand{\zti}{z_t^\i}
\newcommand{\ri}{r^\i}
\renewcommand{\bi}{b^\i}
\newcommand{\ttimesr}{rt}

\begin{document}
\runningauthor{Pierson*, Koh*, Hashimoto*, Koller, Leskovec, Eriksson, and Liang}

%

%

\twocolumn[
\vspace{-12mm}
\aistatstitle{Inferring Multidimensional Rates of Aging from Cross-Sectional Data}
\vspace{-8mm}
\aistatsauthor{Emma Pierson* \And Pang Wei Koh* \And  Tatsunori Hashimoto*}
\aistatsaddress{Stanford and Calico Life Sciences\\emmap1@cs.stanford.edu \And Stanford and Calico Life Sciences \\ pangwei@cs.stanford.edu \And Stanford \\ thashim@cs.stanford.edu } 
\vspace{-4mm}

\aistatsauthor{Daphne Koller \And Jure Leskovec \And Nicholas Eriksson \And Percy Liang}
\aistatsaddress{Calico Life Sciences and Insitro \\ daphne@insitro.com \And Stanford \\ jure@cs.stanford.edu \And Calico Life Sciences and 23andMe  \\ nick.eriksson@gmail.com \And Stanford \\ pliang@cs.stanford.edu}
\vspace{-4mm}
]

\begin{abstract}
\vspace{-4mm}
Modeling how individuals evolve over time is a fundamental problem in the natural and social sciences.
However, existing datasets are often \emph{cross-sectional} with each individual observed only once, making it impossible to apply traditional time-series methods.
Motivated by the study of human aging, we present an interpretable latent-variable model that learns temporal dynamics from cross-sectional data.
Our model represents each individual's features over time as a nonlinear function of a low-dimensional, linearly-evolving latent state.
We prove that when this nonlinear function is constrained to be \emph{order-isomorphic},
the model family is identifiable solely from cross-sectional data provided the distribution of time-independent variation is known.
On the UK Biobank human health dataset, our model reconstructs the observed data while learning interpretable rates of aging associated with diseases, mortality, and aging risk factors.

\end{abstract}
\vspace{-6mm}


\section{Introduction}\label{sec:intro}

Understanding how individuals evolve over time
is an important problem in fields such as
aging \citep{belsky2015quantification},
developmental biology \citep{waddington1940organisers},
cancer biology \citep{nowell1976clonal},
ecology \citep{jonsen2005robust},
and economics \citep{ram1986government}.
However, observing large-scale temporal measurements of individuals is expensive and
sometimes even impossible due to destructive measurements---e.g., in sequencing-based assays \citep{campbell2017uncovering}.
As a result, we often only have \emph{cross-sectional data}---each individual is only measured at one point in time (though different individuals can be measured at different points in time).
From this data,
we wish to learn \emph{longitudinal models} that allow us to make inferences
about how individuals change over time.

This paper is motivated by the problem of studying human aging using data from the UK Biobank,
which contains extensive health data for half a million participants of ages 40--69~\citep{sudlow2015uk}.
As an individual ages, many phenotypes change in correlated ways \citep{mcclearn1997biogerontologic}.
Our goal is to find a low-dimensional latent representation of the phenotype feature space that captures the rates at which individuals change along each dimension as they age.
To be scientifically useful---for instance, in understanding the genetic determinants of aging---these aging dimensions and rates should be interpretable (e.g., grouping phenotypically-related features together) and provably recoverable given some assumptions on the data.

\begin{figure}[b] \vspace{-.4cm} \begin{center} \includegraphics[scale=0.25]{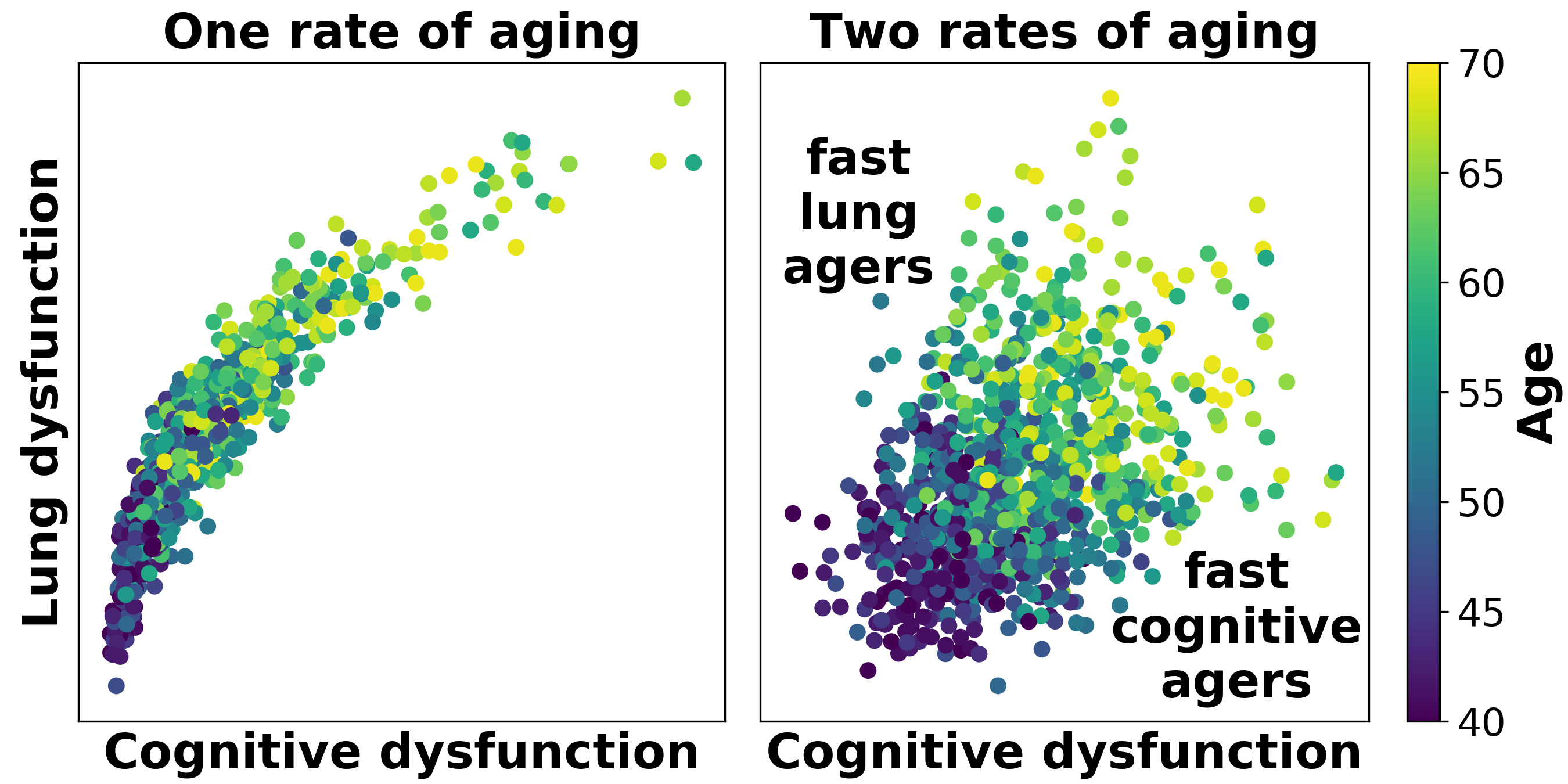} \end{center} \vspace{-.5cm}\caption{\label{fig:pedagogical_figure}A toy example displaying multiple rates of aging (right) which allows individuals to progress rapidly in one aging dimension but not another.} \vspace{-.2cm} \end{figure}

The UK Biobank is unique among health datasets in its breadth and scale.
However, most of its data is cross-sectional:
95\% of its participants are measured at a single time point.
Can we learn how individuals change over time purely from such cross-sectional data?
While impossible in general~\citep{hashimoto2016learning},
this inference has been carried out in restricted settings, e.g., in
single-cell RNA-seq studies~\citep{campbell2017uncovering, trapnell2014dynamics, bendall2014single}.
However, those methods assume that individuals travel along the same
single-dimensional trajectory, whereas human aging is a multi-dimensional process~\citep{mcclearn1997biogerontologic}:
    someone might stay relatively physically fit but experience cognitive decline or vice versa (\reffig{pedagogical_figure}).
Other methods handle multi-dimensional latent processes (e.g., \citet{wang2018learning})
but are concerned with inferring how the population evolves as a whole rather than with individual trajectories, and they do not provide guarantees on the interpretability or identifiability of the latent state.

In this paper, we introduce a method to learn a generative model of a multi-dimensional temporal process from cross-sectional data.
We represent each individual by a low-dimensional latent state comprising
a vector $\ttimesr$ that evolves linearly with time $t$
and a static bias vector $b$ that encodes time-independent variation.
An individual's observations are modeled as a non-linear function of their latent state $\ttimesr$ and $b$.
In the aging context, each component of $r$ captures the age-dependent progression of different groups of phenotypes (e.g., muscle strength vs. cognitive ability),
while $b$ captures age-independent individual variation.

We first study identifiability: under what conditions is it possible to learn the above model from the cross-sectional data that it generates? The key structure we leverage is the
\emph{monotonicity} of the mapping from the time-evolving state $\ttimesr$ to a subset of the observed features.
This captures the intuition that aging is a gradual process where many systems show a generally-monotone decline after the age of 40, e.g., weight \citep{mozaffarian2011changes},
red blood cells \citep{hoffmann2015effect},
lung function \citep{stanojevic2008reference}, and
lean muscle mass \citep{goodpaster2006loss}.
We prove that if the distribution of time-independent variation $b$ is known, a stronger version of monotonicity known as \emph{order isomorphism} implies model identifiability (\refsec{identifiability}).
Our work improves upon known identifiability results~\citep{hashimoto2016learning}
by giving identifiability results for the latent and non-ergodic cases.
We also discuss how to optimize over monotone functions and check for order isomorphisms in our class of models, which is a computationally difficult question of independent interest (\refsec{checking}).

We assess our model using data from a subset of the UK Biobank: specifically, 52 phenotypes measured for more than 250,000 individuals with ages 40--69 (\refsec{experiments-real}). Using this data, we learn an interpretable,
low-dimensional representation of how human phenotypes change with age. This representation accurately reconstructs the observed data and predicts age-related changes in each phenotype.
Through posterior inference on the rate vector $r$, we recover different dimensions of aging corresponding to different coordinates of $r$; these have natural interpretations as belonging to different body systems (e.g., cognitive performance and lung health).
Consistent with biological knowledge, higher inferred rates of aging
are associated with disease, mortality, and known risk factors (e.g., smoking).

\vspace{-2mm}
\section{Model}\label{sec:model}
\vspace{-2mm}

Let $\xti \in \R^d$ be the observed features of the $i$-th individual at time $t$.
A classic approach to modeling temporal progression is to assume that $\xti$ depends linearly
on some scalar, latent measure of progression $\zti \in \R$ (e.g., \citet{klemera2006new}, \citet{levine2012modeling}, and \citet{campbell2017uncovering}).
In the context of human aging, this scalar $\zti$ is often called \emph{biological age}.
We extend this approach in three ways:
we allow $\xti$ to depend non-linearly on $\zti$ and a bias term;
we constrain some components of $\xti$ to depend monotonically on $\zti$;
and we allow $\zti$ to have multiple dimensions.
Specifically, we characterize the $i$-th individual by two latent vectors:
\begin{enumerate}
\item A \emph{rate} vector $\ri \in \R^{k_{\text{r}}}$ that determines how rapidly the $i$-th individual is changing over time.
\item A \emph{bias} vector $\bi \in \R^{k_{\text{b}}}$ that encodes time-independent variation.
\end{enumerate}
Each individual has their own values of $\ri$ and $\bi$ which do not change over time.
For brevity, we will omit the $(i)$ superscript in the sequel unless explicitly comparing individuals.

We model each individual as evolving linearly
in latent space at a rate proportional to $r$, i.e., $z_t = \ttimesr$,
and we model $x_t$ as the sum of a time-dependent term $f(\ttimesr)$ and time-independent term $g(b)$:
\begin{align}
x_t = f(\ttimesr) + g(b) + \epsilon.
\end{align}
Here, $f\colon \R^{k_{\text{r}}} \to \R^{d}$
is a non-linear function capturing time-dependent variation;
$g\colon \R^{k_{\text{b}}} \to \R^d$ is a non-linear function capturing time-independent variation;
and $\epsilon$ is measurement noise sampled i.i.d.\ at each time point. ${k_{\text{r}}}$ and ${k_{\text{b}}}$ are model hyperparameters. We assume that $r$, $b$, and $\epsilon$ are independently drawn from known priors,
and that the rates $r$ are always positive.

\paragraph{Interpretation.} We interpret $z_t = \ttimesr$ as an individual's `biological age'.
In contrast to previous work, $z_t$ and $r$ are vector-valued quantities, capturing the intuition that aging is a multi-dimensional process (as discussed in \refsec{intro}).
The function $f$ links the biological age $z_t$ with the observed features (phenotypes) $x_t$.
The rate $r$ describes how quickly an individual ages along each latent dimension and differs between individuals, since different individuals experience age-related decline at different rates \citep{mcclearn1997biogerontologic}.
The bias $b$ captures non-age-related variation like intrinsic differences in height, and also differs between individuals.

\paragraph{Monotonicity.}
To ensure that the model is identifiable from cross-sectional data, we assume that some coordinates of $f$ are \emph{monotone}.
Roughly speaking, as $t$ increases, those coordinates of $f(\ttimesr)$ also increase on average;
we defer a precise definition to \refsec{identifiability}.
The monotonicity of $f$ is a reasonable assumption in the setting of human aging, as many features vary monotonically with age after the age of 40 \citep{mozaffarian2011changes, hoffmann2015effect, stanojevic2008reference, goodpaster2006loss}.
Monotonicity does not imply that, e.g., an individual's strength has to strictly decrease with age (due to $\epsilon$) or that an older individual is always weaker than a younger one (because $g$ allows for age-independent variation between people).

For simplicity, we assume that the monotone phenotypes are known in advance.
To streamline notation, we define $f$ to be monotone and have all non-monotone features modeled by some other unconstrained $\tilde{f}$, i.e.,
$x_t = [f(\ttimesr); \tilde{f}(\ttimesr)] + g(b) + \epsilon$.

\paragraph{Learning.}
Our goal is to estimate $f$, $\tilde{f}$, $g$, and $\epsilon$ from cross-sectional data
$\{(t^{(i)}, x_{t^{(i)}}^{(i)})\}_{i=1}^n$.
We parametrize the functions with neural networks and use a variational autoencoder
to optimize a standard lower bound on the likelihood of the observed data \citep{kingma2014variational};
see \refsec{implementation} for more details.

\section{Identifiability}\label{sec:identifiability}

\newcommand{\fmono}{f}
\newcommand{\xmono}{x}
\newcommand{\fnonmono}{h}
\newcommand{\xnonmono}{x}

We first study the basic question of identifiability: is it possible to recover $f$ (and thereby estimate temporal dynamics and rates of aging $r$) from cross-sectional data that is generated by $f$?  In other words, do different $f$ give rise to different observed data?

Without loss of generality, we make two simplifications in our analysis.
First, we only consider features $x$ that correspond to the monotone $f$,
and disregard those which correspond to $\tilde{f}$;
if the model is well-specified and can be identified just by considering $f$,
then it will remain identifiable when additionally considering the non-monotone part $\tilde{f}$.
Second, we consider a single noise term $\epsilon' \eqdef g(b) + \epsilon$ which combines age independent variation $g(b)$ and the measurement noise $\epsilon$, since this does not affect the rate of aging.\footnote{
  In \refsec{model}, we separate $g(b)$ and $\epsilon$,
  since it might be possible to estimate these quantities separately based on prior literature or a small amount of longitudinal data.
}
Together, these give the simplified model
\begin{align}\label{eq:simplemodel}
x_t &= f(\ttimesr) + \epsilon',
\end{align}
where $x_t \in \R^d$ are the observed features, $r \in \R^{k_r}$ is the rate vector, and $t \in \R_+$ and $\epsilon' \in \R$ are scalars.
\newcommand{\disteq}{\overset{d}{=}}
If $f$ is a general differentiable function without any monotonicity constraints,
there exist functions that are unidentifiable from observations of the distribution of $x_t$.
As an example,
consider $\epsilon' = 0$ and $r \sim \text{lognormal}(0,1)$. Let $M$ be any matrix that preserves the all-ones vector $\bone$ (i.e., $M \bone = \bone$) and is an orthogonal transform on the orthogonal subspace to $\bone$.
Since $\log(\ttimesr) \sim \sN(\log t \bone, 1)$,
$M \log(\ttimesr) \disteq \log(\ttimesr)$ due to the rotational invariance of the Gaussian (where $\disteq$ means equality in distribution). This implies that
$f(\exp(M \log(\ttimesr)))\disteq f(\ttimesr)\disteq x_t$.
Since $f(\exp(M \log(\cdot)))$ and $f(\cdot)$ have the same observed distribution, they are indistinguishable from each other.

Therefore, we need to make additional assumptions on $f$ to ensure identifiability.
Here, we will show that $f$ is identifiable up to permutation whenever the distribution of $\epsilon'$ is known and
both $f$ and $f^{-1}$ are monotone---that is, $f$ is an \emph{order isomorphism}.

\begin{defn}\label{defn:monotone}
A function $f$ is \emph{monotone} if
$u \preceq v \implies f(u) \preceq f(v)$ for all $u, v \in \text{dom}(f)$,
where ordering is taken with respect to the positive orthant
(i.e., $u \preceq v$ means $u_i \leq v_i$ for all $i$).
\end{defn}
\begin{defn}\label{defn:order-iso}
An injective function $f$ is an \emph{order isomorphism} if $f$ and $f^{-1}$ restricted to the image of $f$ are both monotone, that is,
$u \preceq v \iff f(u) \preceq f(v)$.\footnote{We deviate from standard nomenclature, where order-isomorphic $f$ are defined as bijections, by letting $f$ be injective and considering the restriction of $f$ to its image.}
\end{defn}

\subsection{Noiseless setting ($\epsilon'=0$)}

We begin by considering the case where $\epsilon'=0$.
Our main identifiability result is the following:

\begin{proposition}\label{prop:identifiability}
  Let $x_t$ and $\ttimesr$ be the random variables defined in  \eqref{eq:simplemodel}.
  If $f_1$ and $f_2$ and their inverses are twice continuously differentiable and are order-isomorphic functions such that
 $f_1(\ttimesr) \disteq f_2(\ttimesr) \disteq x_t$ for some $t>0$, then $f_1$ and $f_2$ are identical up to permutation.\footnote{We define $f_1$ and $f_2$ as identical up to permutation if there exists a permutation matrix $P$ such that $f_1(Pv) = f_2(v)$.}
\end{proposition}
We defer full proofs to Appendix \ref{sec:proofs}, but provide a short sketch here. The proof consists of two parts: we first show that all bijective order isomorphisms are permutations followed by component-wise monotone transforms. Then we show that any two maps $f_1$ and $f_2$ matching the observed data must be identical up to permutation.
\begin{lemma}\label{lem:iso-perm-coord}
  If $\diffmap\colon \mathbb{R}^{k_{\text{r}}} \to \mathbb{R}^{k_{\text{r}}}$ is twice continuously differentiable and an order isomorphism, $\diffmap$ must be expressible as a permutation followed by a component-wise monotone transform.
\end{lemma}
We then consider the difference map $\diffmap \eqdef f_2^{-1} \circ f_1$, which maps the latent state implied by $f_1$ to that of $f_2$.
$\diffmap$ is also an order isomorphism, so by \reflem{iso-perm-coord} it is the composition of a permutation and monotone map. Since $f_1(\ttimesr) \disteq f_2(\ttimesr) \disteq x_t$, $\diffmap$ is measure preserving for $\ttimesr$. As the only monotone measure preserving map is the identity, $\diffmap$ must be a permutation.

\subsection{Noisy setting ($\epsilon' \neq 0$)}

Identifiability in the noisy setting is more challenging. If the noise distribution $\epsilon'$ is known, we can reduce the noisy setting to the noiseless setting by first taking the observed distribution of $x_t$ and then deconvolving $\epsilon'$. This gives us the distribution of $f(x_t)$, to which we can apply Proposition \ref{prop:identifiability}. The uniqueness of this procedure follows from the uniqueness of Fourier transforms and inverses over $L^1$ functions \citep{stein2011fourier}.
This corresponds to the setting where we can characterize the distribution of the time-independent variation $g(b)$ and the measurement noise $\epsilon$, either through prior knowledge or measurement
(e.g., in a controlled setting where we observe the starting point $x$ of all individuals).
Importantly, we do not need to know the exact value of $b$ and $\epsilon$ for any individual, just their distributions.

If the noise distribution $\epsilon'$ is unknown, then the characterization we provide here no longer holds, and we cannot simply deconvolve the noise.
Nevertheless, we conjecture that the strong structure induced by monotonicity is sufficient for identifiability, and in simulations we are able to recover known ground-truth parameters (\refsec{experiments-synth}).

\vspace{-2mm}
\section{Learning order isomorphisms}\label{sec:checking}
\vspace{-2mm}
\newcommand{\monoclass}{\mathcal{M}}

Our identifiability results suggest that
we should optimize for $f$ within the class of order isomorphisms.
However, that optimization is difficult in practice, as it requires
constraints on $f^{-1}$ that hold over the entire image of $f$.
Instead, we take the following approach:
\begin{enumerate}
    \item We relax the order isomorphism constraint and optimize for $f$ within a class of monotone transformations $\monoclass$ that have a particular parametrization.
    \item We check, post-hoc, if the learned $f \in \monoclass$ is approximately order-isomorphic. (In real-world optimization settings, $f$ will not be exactly order-isomorphic for reasons we discuss below.)
    While not all functions in $\monoclass$ are order-isomorphic,
    we choose $\monoclass$ such that we can quickly verify if a given $f \in \monoclass$ is approximately order-isomorphic.
\end{enumerate}
While we do not have any prior expectation that the learned $f$ would be order-isomorphic,
surprisingly, we find in our experiments (Section \ref{sec:implementation}) that we do in fact learn an approximately order-isomorphic $f \in \monoclass$.
This suggests that we do not lose any representational power by moving from monotone functions to order-isomorphic functions, and that the assumption of order isomorphism (on top on monotonicity) is reasonable.

We choose $\monoclass$ to be the set of functions that can be written as
$f \colon \mathbb{R}^k \to \mathbb{R}^d = s_2 \circ \lineartransformation \circ s_1$,
where $s_1 \colon \mathbb{R}^k \to \mathbb{R}^k$ and $s_2 \colon \mathbb{R}^d \to \mathbb{R}^d$ are continuous, component-wise monotone transformations,\footnote{%
$s$ is a component-wise transformation if it acts separately on each component of its input, i.e., $s(v) = [s_1(v_1), s_2(v_2), \ldots, s_k(v_k)]$.
} and $\lineartransformation \colon \mathbb{R}^k \to \mathbb{R}^d$ is a linear transform.
All $f \in \monoclass$ are monotone by construction, due to the compositionality of monotone functions.

The following results show that we can check if some $f \in \monoclass$ is order-isomorphic, i.e., $f^{-1}$ is also monotone, by examining only
the linear transform $\lineartransformation$:
\begin{lemma}\label{lem:injective-linear}
  Let $\lineartransformation(v) = Av$ be a linear transform, where $A \in \mathbb{R}^{d \times k}$.
  If we can write $A = P\left[\begin{matrix} B \\ C \end{matrix}\right]$ where $P$ is a permutation matrix, $B$ is a non-negative monomial matrix,\footnote{
  A monomial matrix is a square matrix in which each row and each column has only one non-zero element. In other words, it is like a permutation matrix, except that the non-zero elements can be arbitrary.} and $C$ is a non-negative matrix, then $a(\cdot)$ is an order isomorphism.
\end{lemma}
\begin{proposition}\label{prop:iso-checking}
Let $f \colon \mathbb{R}^k \to \mathbb{R}^d = s_2 \circ \lineartransformation \circ s_1$,
where $s_1 \colon \mathbb{R}^k \to \mathbb{R}^k$ and $s_2 \colon \mathbb{R}^d \to \mathbb{R}^d$ are continuous, component-wise monotone transformations,
and $\lineartransformation \colon \mathbb{R}^k \to \mathbb{R}^d$ is a linear transform.
If $a$ satisfies \reflem{injective-linear},
then $f$ is an order isomorphism.
\end{proposition}
See Appendix \ref{sec:proofs-checking} for proofs.
Correspondingly, during training, we can restrict $f$ to the form $s_2 \circ \lineartransformation \circ s_1$, where $s_1$ and $s_2$ are component-wise monotone transforms and $\lineartransformation$ is a linear transformation parametrized by $A$, a non-negative matrix.
To check if the learned $f$ is order-isomorphic, \refprop{iso-checking} tells us that it suffices to check if $A$ satisfies the conditions of \reflem{injective-linear}.
Equivalently, each column of $A$ must have a non-zero element
in a row where every other column has a zero.

\paragraph{Implementation.}
The results above apply to linear transforms $a$ with pre- and post-transformations $s_1$ and $s_2$. In our experiments (\refsec{implementation}), however, we found that using a single monotone component-wise transform ($f = s \circ \lineartransformation$) did not significantly harm performance. To help interpretability, we thus only use a single component-wise transform $s$.
We parametrized $s$ as a polynomial with non-negative coefficients; this can be swapped for other differentiable parametrizations of monotone functions \citep{gupta2016monotonic}.
This $f$ can be optimized during training by applying gradient descent to $A$ and the coefficients of $s$.

In our fitted model (\refsec{experiments-real}), the learned $A$ was close to satisfying \reflem{injective-linear}:
each column $j$ contained at least one row $i$ where $A_{ij} \gg A_{ik}$ for all $k \neq j$ (specifically, $A_{ij} > 50A_{ik}$).
Thus, learning a monotone $f$ gave us an approximately monotone $f^{-1}$ without further constraints.
This empirical finding was surprising to us and warrants future study, since learning an order-isomorphic $f$ would otherwise be computationally hard.

\section{Experimental setup}\label{sec:implementation}

\vspace{-2mm}

\paragraph{Data processing.} Appendix \ref{sec:biobank} describes the full processing procedure. In brief,
we selected features that were measured for a large proportion of participants,
resulting in 52 phenotypes which we categorized by visual inspection into monotone (45/52)
and non-monotone phenotypes (7/52).
For convenience, we pre-processed the monotone phenotypes to all be monotone increasing with age
by negating them if necessary.
We use a train/development set of 213,510 individuals
with measurements at a single time point,
and report all results on a separate test set of 53,174 individuals not used in model development or selection.
We also have longitudinal data from a single follow-up visit for an additional 8,470 individuals.

\vspace{-2mm}

\paragraph{Model details.}\label{sec:model_details}
We used a variational autoencoder to learn and perform inference in our model~\citep{kingma2014variational}. \reffig{model_diagram} illustrates the model architecture.
We parametrize the monotone function $f = s \circ \lineartransformation$
as the composition of a monotone elementwise transformation $s \colon \R^{d'} \to \R^{d'}$
with a monotone linear transform $\lineartransformation \colon \R^{k_{\text{r}}} \to \R^{d'}$.
As described in \refsec{checking}, we parametrized the linear transformation $\lineartransformation$ using a matrix $A$ constrained to have non-negative entries,
and implemented each component $s_i(v) \colon \R_+ \to \R_+$ of $s$
as the sum of positive powers of $v \in \R_+$ with non-negative coefficients
$s_i(v) = \sum_{p_j \in S} w_{j} v^{p_{ij}}$,
where $w_{ij}$ are learned non-negative weights, and $S$ is a hyperparameter.
We verified that the learned model's $A$ matrix 
can be row-permuted into a combination of an approximately monomial matrix and positive matrix,
indicating that we learned an $f$ that was order-isomorphic (\refsec{checking}). For full details, see Appendix \ref{sec:model-appendix}. Our model implementation is publicly available: \url{https://github.com/epierson9/multiphenotype_methods}.

\paragraph{Hyperparameter selection.}

\vspace{-2mm}

We selected all hyperparameters other than the size of the latent states $k_{\text{r}}$ and $k_{\text{b}}$ (e.g., network architecture and the set of polynomials $S$) through random search evaluated on a development set (Appendix \ref{sec:model-appendix}).
Increasing $k_{\text{r}}$ and $k_{\text{b}}$ gives the model more representational power; indeed, test ELBO increased uniformly with increasing $k_{\text{b}}$ and $k_{\text{r}}$ in the range we tested
($k_{\text{b}} + k_{\text{r}} \leq 20$).
We chose $k_{\text{b}} = 10$ and $k_{\text{r}} = 5$ to balance modeling accuracy with dimensionality reduction for interpretability,
since the test ELBO begins to level off at $k_{\text{r}} = 5$;
we chose a higher $k_{\text{b}}$ since we are not concerned with compressing the time-independent variation.
Our results were similar with other values of $k_{\text{r}}$ and $k_{\text{b}}$ (Appendix \ref{sec:robustness_checks}).

\vspace{-2mm}

\section{Results on synthetic data}
\label{sec:experiments-synth}

\vspace{-2mm}

\newcommand{\rtrue}{\ensuremath{r_{\text{true}}}}
\newcommand{\rfitted}{\ensuremath{r_{\text{fitted}}}}

To check if we could correctly recover the rates of aging $r$ in the well-specified setting, we generated synthetic data from a model and tried to recover the model parameters from that data.

We measured the quality of recovery by comparing the correlation between ground truth rates of aging $\rtrue$ and predicted individual rates of aging $\rfitted$. To generate realistic synthetic data, we fit the model described in \refsec{implementation} to data from the UK Biobank
and then sampled from it (using the stated priors on $r$, $b$, and $\epsilon$).
We verified that this synthetic data matched the properties of UK Biobank data, such as the age trends
for each feature (\refsec{experiments-real} provides details). We ran this check for models with different values of $k_{\text{r}} = 1, 2, \dots, 10$
and found good concordance across all values of $k_{\text{r}}$:
the mean correlation between $\rfitted$ and $\rtrue$ was $0.91$ (averaged across values of $k_{\text{r}}$ and dimensions of $r$),
and the slopes of the regressions of $\rfitted$ on $\rtrue$ were very close to one (mean absolute difference from 1 of 0.09), indicating good calibration.

These results suggest that if the model is well-specified, then it is identifiable even though the distribution of $g(b) + \epsilon$ is not known \textit{a priori}. Moreover, our training procedure is able to recover the ground truth parameters quite closely.

\vspace{-2mm}
\section{Results on UK Biobank data}\label{sec:experiments-real}
\vspace{-2mm}

We first verify that our model fits the data (\refsec{model_accuracy}), before showing, as our main result, that it yields interpretable and biologically plausible rates of aging (\refsec{model_interpretation}). We compare to four baselines: principal components analysis (PCA); mixed-criterion PCA (mcPCA) \citep{bair2006prediction}; contrastive PCA (cPCA) \citep{abid2018exploring}; and our model with the monotone constraints removed and the same hyperparameter settings. We evaluate PCA, mcPCA, and cPCA using the same number of latent states as the original model ($k_{\text{r}} + k_{\text{b}} = 15$); Appendix \ref{sec:baselines-appendix} provides full implementation details. (We discuss other potential baselines, and why they cannot be applied in our setting, in \refsec{related}).

\subsection{Reconstruction and extrapolation}\label{sec:model_accuracy}

We first show that our model can \emph{reconstruct} each individual's features  from their low-dimensional latent state and \emph{extrapolate} to future timepoints.
The goal of these evaluations is not to demonstrate state-of-the-art predictive performance; rather, we want to verify that our model accurately reconstructs individual datapoints and captures aging trends.

\paragraph{Reconstruction.} We assessed whether our model was able to reconstruct observed datapoints from their latent space projections. Given an observation $(t, x_t)$,
we computed the approximate posterior mean of the latent variables $(\hat{r}, \hat{b})$ using the encoder,
and compared $x_t$ against the reconstructed posterior mean of $x_t$ given $(\hat{r}, \hat{b})$.
On a held-out test set, reconstruction was largely accurate,
with a mean correlation between true and reconstructed feature values of 0.88 ( \reffig{fig:scatter_plots}). The other baselines performed similarly: PCA, mcPCA, and cPCA did slightly worse,
with mean correlations of 0.86, 0.86, and 0.84 respectively.
The non-monotone model did slightly better (mean correlation 0.89); the small difference demonstrates that our monotone assumption does not undermine model fit.

\paragraph{Extrapolation to future timepoints.} To assess how accurately the model captures the dynamics of aging, we evaluate its ability to `fast-forward' people through time: that is, to predict their phenotype $x_{t_1}$ at a future age $t_1$ given their current phenotype $x_{t_0}$ at age $t_0$.
As above, we compute the posterior means $(\hat{r}, \hat{b})$ using $x_{t_0}$ and $t_0$; we then predict $x_{t_1} = f(t_1 \ \hat{r}) + g(\hat{b})$. We do not compare to PCA, mcPCA, and cPCA on this task because they do not provide dynamics models, making it impossible to perform fast-forwarding.

\vspace{-.2cm}
\begin{figure}[ht] \begin{center} \includegraphics[scale=0.18]{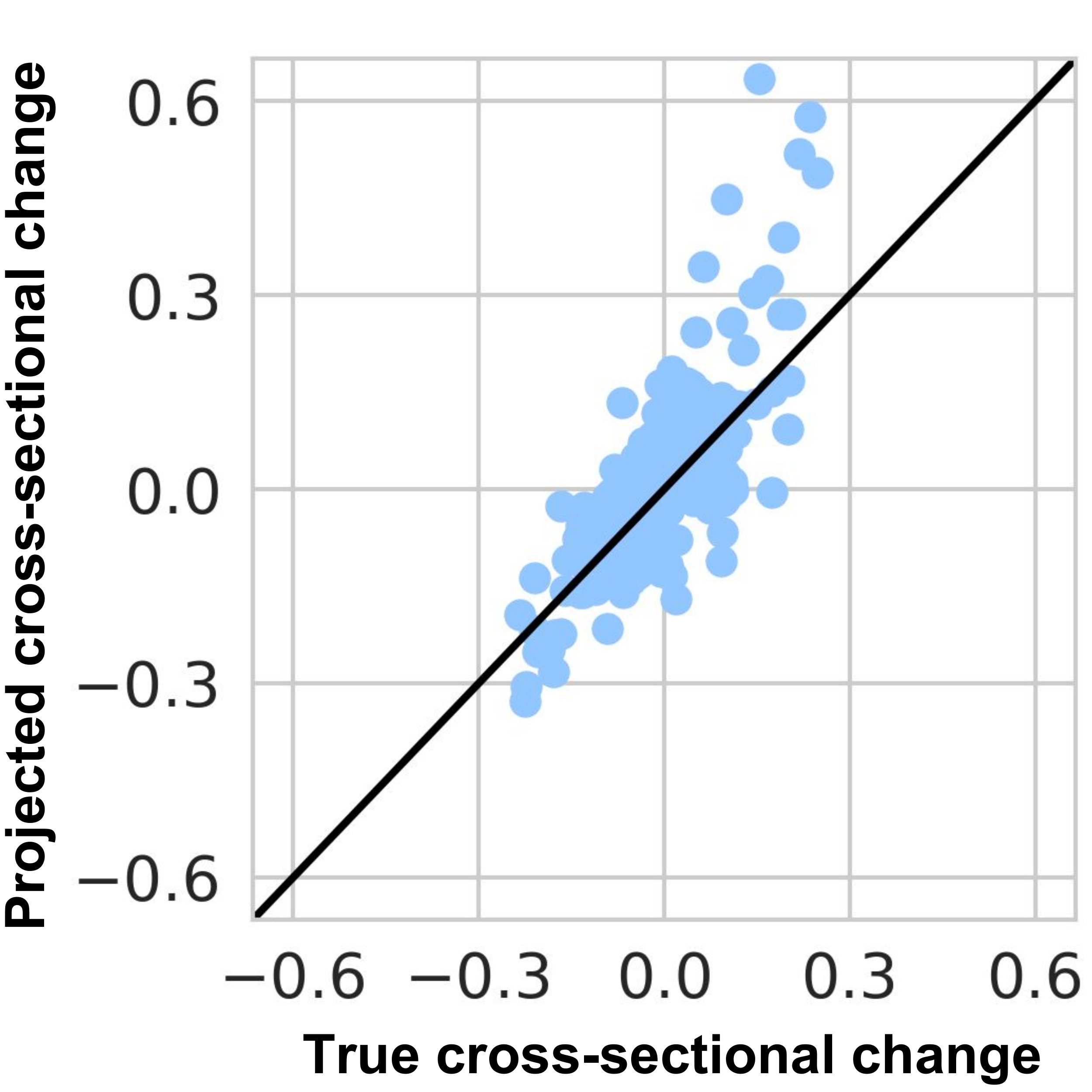} \end{center} \vspace{-.5cm}\caption{\label{fig:xs_fast_forwarding}True and predicted changes are well-correlated ($r = 0.77$) in cross-sectional data when fast-forwarding 5 years. Each point represents difference in one feature for one 5-year age bin: e.g., difference in heart rate between 40--45 and 45--50 year olds.} \end{figure}\vspace{-.3cm}

We assess the accuracy of fast-forwarding on both cross-sectional and longitudinal data.
On cross-sectional data, we do not have follow-up data $x_{t_1}$ for each person,
so we evaluate the model by \emph{age group}:
for example, we fast-forward all 40--45 year olds by 5 years,
compute how much the model predicts each feature will change on average,
and compare that to the true average feature change between 40--45 year olds and 45--50 year olds. (Although we bucket age in this analysis to reduce noise, the model uses the person's exact age.)
Predictions are highly correlated with the true values ($r=0.77$ for 5-year follow-ups, \reffig{xs_fast_forwarding}; $r=0.88$ for 10 years, and $0.94$ for 15 years. This is similar to the performance of the non-monotone baseline, which achieves correlations of $0.77, 0.90$, and $0.96$ for 5, 10, and 15 years.

On longitudinal data, we observe both $x_{t_0}$ and $x_{t_1}$ for a single person,
and can therefore use reconstruction accuracy of $x_{t_1}$ as a metric. This task is difficult because longitudinal follow-up times are very short in our dataset (2--6 years),
so aging-related changes may be swamped by the inherent noise in the task and sampling biases in the longitudinal cohort.
We compare to three additional baselines on this task: predicting no change, $x_{t_1}=x_{t_0}$;
reconstructing $x_{t_0}$ without fast-forwarding, $x_{t_1} = f(t_0 \ \hat{r}) + g(\hat{b})$; and
fast forwarding according to the average rate of change in the cross-sectional data. Our evaluation metric is the fraction of people for which our model yields lower reconstruction error than each baseline.\footnote{We use this metric over the mean error because the noise in the data is large relative to aging-related change, so the mean improvement for a particular individual will be small even if one method consistently yields better predictions.} For follow-up times long enough to allow for substantial age-related change (${\ge5}$ years),
our model predicts $x_{t_1}$ more accurately than all three benchmarks on most individuals (Table \ref{tab:longitudinal_extrapolation}, top row), and performs comparably, though slightly worse, than the non-monotone model (Table \ref{tab:longitudinal_extrapolation}, bottom row).
Appendix \ref{sec:longitudinal_extension_experiments} describes a natural extension of our model which allows both longitudinal and cross-sectional data to be used in model fitting, which significantly improves performance on this task.

\vspace{-.4cm}
\begin{table}[h]
  \caption{\% of people for which the rate-of-aging models predict $x_{t_1}$ more accurately than do benchmarks.} \label{tab:longitudinal_extrapolation}
\vspace{-.2cm}
\begin{center}
\small
\begin{tabular}{|l|p{1cm}|p{1.8cm}|p{1.0cm}|}
\hline
Benchmark methods: & $x_{t_0}$ & Recons. $x_{t_0}$ & avg $\Delta$ \\
\hline
Monotone     & 66\%  & 61\% & 60\% \\ \hline
Non-monotone   & 71\% & 63\% & 65\% \\ \hline
\end{tabular}
\end{center}
\end{table}
\vspace{-.3cm}

The results above show that our model reconstructs the observed data slightly more accurately than linear methods (PCA, cPCA, and mcPCA) while providing an accurate dynamics model, which these linear methods do not.
Moreover, the monotonicity assumption does not hurt our model's performance too much.

\subsection{Model interpretation}\label{sec:model_interpretation}

\begin{figure*}[ht] \begin{center} \includegraphics[scale=0.28]{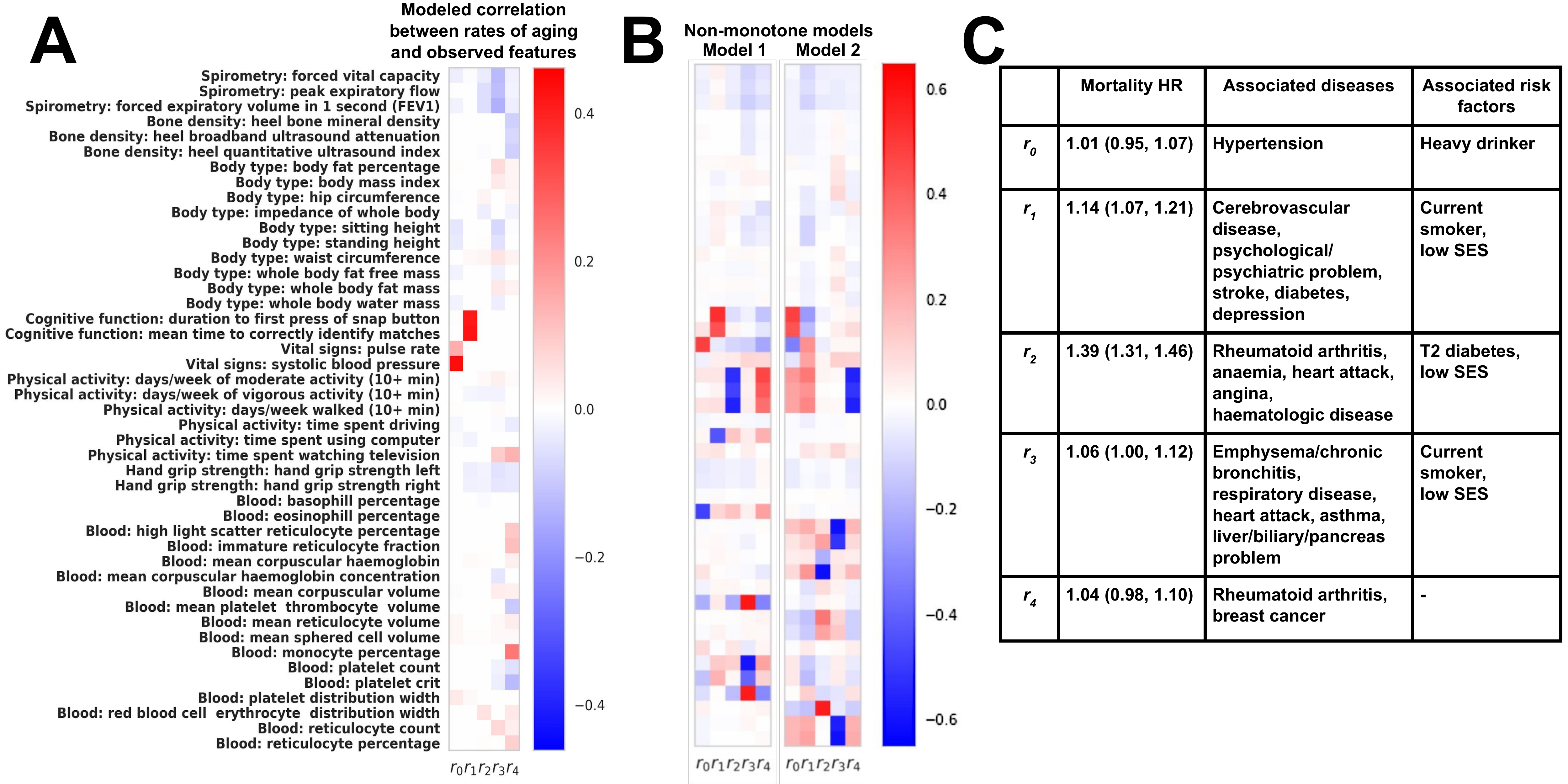} \end{center} \vspace{-.5cm}\caption{\textbf{A}: Model-learned correlations between rates of aging $r$ (columns) and observed features $x$ (rows) for monotone features. Each cell displays the correlation between one rate of aging and one observed feature in data sampled from the model. Learned rates of aging are sparse and stable across runs and hyperparameter settings; in contrast, without monotone constraints (\textbf{B}) the rates of aging are not sparse or stable. \textbf{C}: Associations with mortality, diseases, and rate-of-aging risk factors. Mortality HR is the hazard ratio for a one-std-dev increase in rate of aging in a Cox proportional hazards model. The final two columns list the diseases and risk factors most strongly positively associated with higher rates of aging. We list up to five significant ($p < 0.05$ after Bonferroni correction) and strong (effect size $>$ 1\% increase in the rate of aging) associations for each rate.\label{fig:monotonic_feature_interpretations}}\vspace{-.4cm} \end{figure*}

Our main experimental result is that we obtain interpretable rates of aging from our monotone model.
In particular, we found that enforcing monotonicity in $f$ encouraged sparsity. To interpret the rates of aging $r$, we simply associated each component of $r$ with the sparse set of features that it correlated with (\reffig{monotonic_feature_interpretations}A). These rates were more interpretable than those learned by the four baselines: the model without monotone constraints and PCA, cPCA, and mcPCA. Without the monotone constraints (\reffig{monotonic_feature_interpretations}B) the rates are not associated with sparse sets of features and are less interpretable. Further, because the rates of aging $r$ in the non-monotone model can be rotated without affecting model fit, the model is unstable, learning different rates $r^\i$ for the same individual $i$ when trained with different random seeds. \reffig{monotonic_feature_interpretations}B shows that the top two non-monotone models (by test ELBO) learn very different $r$'s. To compare two models, we defined $\corrbetweenmodels$ as the correlation between the $r^\i$'s learned by the two models, averaged over each component of $r$, and maximized over permutations of components. For the top two non-monotone models,  $\corrbetweenmodels$ was only 0.44, vs. $\corrbetweenmodels=0.94$ for the top two monotone models. The monotone model was also stable over changes to the number of non-monotone features, random subsets of the training data, and the dimensions $k_r$ and $k_b$ of the time-dependent and bias latent variables (Appendix \ref{sec:robustness_checks}). This demonstrates empirically that, consistent with our theoretical results, monotonicity is essential to learning stable and interpretable rates of aging.

Similarly, the latent factors learned by the three linear baselines (PCA, cPCA, and mcPCA) were difficult to interpret because they did not clearly distinguish between time-dependent and time-independent variation and were not sparse (Appendix \reffig{fig:linear_baseline_loadings}). For example, the first principal component learned by PCA was most strongly associated with body type (e.g., height and weight), but this mostly captures the variation in body type within age groups, and is only weakly correlated ($\rho = 0.13$) with age. None of the top 5 principal components learned by any of the three methods had an absolute correlation with age of greater than 0.3.

To assess the biological plausibility of our learned rates of aging, we examined associations between each rate of aging and three external sets of covariates not used in fitting the model: mortality; 91 diseases; and 5 risk factors which are known to accelerate aging processes, such as being a current smoker. We show these associations in \reffig{monotonic_feature_interpretations}C (see Appendix \ref{sec:biobank} for details). Rates of aging were positively associated with all three sets of covariates: of the 88 statistically significant associations with diseases ($p < 0.05$ after Bonferroni correction), 78\% were positive; 73\% of the 15 statistically associations with risk factors were positive, and all associations with mortality were positive although---interestingly---to widely varying degrees.

Based on these associations, we interpret the rates of aging $r$ as follows: $r_0$, a `blood pressure rate of aging', associates with blood pressure;
$r_1$, a `cognitive rate of aging', associates with the two cognitive function phenotypes and with cognitive diseases (e.g., psychiatric problems and strokes);
$r_2$ associates with heart conditions including heart attacks and angina
 and is the most strongly associated with mortality;
$r_3$, a `lung rate of aging', associates with pulmonary function and lung diseases (e.g., bronchitis and asthma), and is elevated in smokers;
and $r_4$, a `blood and bone rate of aging', correlates with blood phenotypes (e.g., monocyte percentage) and rheumatoid arthritis, an autoimmune disorder associated with changes in monocyte and platelet levels \citep{milovanovic2004relationships,rossol2012cd14brightcd16}. Interestingly, $r_4$ also correlates with the bone density phenotypes, a direction for future study.

\vspace{-2mm}

\section{Related work}\label{sec:related}

\vspace{-2mm}

\paragraph{Biological age.}
In our work, we interpreted the vector $\ttimesr$ as the `biological age' of an individual.
The notion of biological age as a measurable quantity that tracks chronological age on average
but captures an individual's `true age'
dates back 50 years \citep{comfort1969test}.
It is common to regress chronological age against a set of phenotypes and
call the predicted quantity biological age
\citep{furukawa1975assessment, borkan1980assessment, klemera2006new, levine2012modeling, horvath2013dna, chen2016dna, putin2016deep}.
These methods estimate a single-dimensional biological age
and do not allow for longitudinal inferences.
\citet{belsky2015quantification} estimates a single-dimensional rate of aging, but requires longitudinal data.

\paragraph{Pseudotime methods in molecular biology.} These methods order biological samples (for example, microarray data
\citep{magwene2003reconstructing, gupta2008extracting} or RNA-seq data
\citep{reid2016pseudotime, kumar2017understanding}) using their gene expression levels; the imputed temporal order is referred to as \emph{pseudotime}.
These methods typically use either some form of minimum spanning tree \citep{qiu2011discovering, trapnell2014dynamics, bendall2014single} or a Bayesian approach \citep{campbell2017uncovering, aijo2014methods}, under the assumption of a single-dimensional temporal trajectory
(with discrete branching points).
\citet{gupta2008extracting} showed recoverability of such methods under similar assumptions.

\paragraph{Dimensionality reduction.}
Others have studied aging on cross-sectional data using dimensionality reduction (DR) methods
such as PCA \citep{nakamura1988assessment} and
factor analysis \citep{macdonald2004biological}, using the first factor as the `aging dimension'.
These methods do not explicitly take temporal information into account,
and therefore do not cleanly factor out time-dependent changes from time-independent changes.
DR methods specific to time-series data,
such as functional PCA, have been used to study clinically-relevant changes over time
\citep{di2009multilevel, greven2011longitudinal} but require longitudinal data.

\paragraph{Recovery of individual dynamics from cross-sectional data.}

Recovering the behavior of individuals from population data has been studied as `ecological inference' \citep{king2013solution} or `repeated cross-section' analysis
\citep{moffitt1993identification, collado1997estimating, kalbfleisch1984least, van1983estimation, hawkins2000estimating, bernstein2016consistently}. These works focus on models without latent variables and are restricted to linear or discrete time-series. \citet{hashimoto2016learning} consider learning dynamics from cross-sectional data in more general settings, but do not consider latent variable inference; moreover, their method relies on observing nearly-stationary data, which is inapplicable to our setting.
\citet{wang2018learning} uses a latent-variable model to infer population evolution, and can also be applied to modeling individuals. However, because their main goal is population dynamics, their latent variables are not designed to be interpretable or identifiable.

\paragraph{Monotone function learning.}
The task of learning partial monotone functions has been well-studied \citep{gupta2016monotonic, daniels2010monotone, qu2011generalized, you2017deep}.
The difficulty in applying these to our setting
is that we need $f$ to have a specific parametric form for efficient order-isomorphism checking (\refsec{checking}), which these methods do not satisfy. It is an open question if these methods can be adapted to learning order isomorphisms.

\vspace{-2mm}
\section{Discussion}
\vspace{-2mm}

We have presented a method to learn, from cross-sectional data, a low-dimensional latent representation of how people change as they age.
Empirically, this representation is interpretable and biologically plausible, allowing us to
infer an individual's rates of aging along each dimension of the latent space.
Theoretically, we leverage the order isomorphism of the mapping between the latent space and the observed phenotypes to show that our model family is identifiable.
To learn an order-isomorphic mapping---which is computationally intractable in general---we introduce a parametric mapping that is easily verifiable as order-isomorphic,
and show through experiments that this parametrization automatically learns an order isomorphism on our data.

Our model opens up many directions for future work. We could extend it to incorporate more complexities of real-world data, including survivorship bias \citep{fry2017comparison, louis1986explaining} or discontinuous changes in latent state (e.g, damage caused by a heart attack).
Powerful previous ideas in latent variable models---for example, discrete latent variables \citep{jang2017categorical, maddison2016concrete} that capture phenomena like sex differences---could be used to relax the model's parametric assumptions. Incorporating genetic information also represents a promising direction for future work. For example, genotype information could be used to learn rates of aging with a stronger genetic basis.

We anticipate that our learned rates of aging will be useful in downstream tasks like genome-wide association studies, where combining multiple phenotypes can increase power \citep{o2012multiphen}.
Finally, we hope that our model, by offering an interpretable multi-dimensional characterization of temporal progression, can be applied to longitudinal inference in other domains, like single-cell analysis and disease progression.

\paragraph{Acknowledgments} We thank Zhenghao Chen, Jean Feng, Adam Freund,
Noah Goodman, Mitchell Gordon, Steve Meadows, Baharan Mirzasoleiman, Chris Olah, Nat Roth, Camilo Ruiz, Christopher Yau,
and the Calico UK Biobank team for helpful discussion. EP was supported by the Hertz and NDSEG Fellowships and PWK was supported by the Facebook Fellowship.

\bibliography{refdb/all}

\begin{thebibliography}{62}
\providecommand{\natexlab}[1]{#1}
\providecommand{\url}[1]{\texttt{#1}}
\expandafter\ifx\csname urlstyle\endcsname\relax
  \providecommand{\doi}[1]{doi: #1}\else
  \providecommand{\doi}{doi: \begingroup \urlstyle{rm}\Url}\fi

\bibitem[Abid et~al.(2018)Abid, Zhang, Bagaria, and Zou]{abid2018exploring}
A.~Abid, M.~J. Zhang, V.~K. Bagaria, and J.~Zou.
\newblock Exploring patterns enriched in a dataset with contrastive principal
  component analysis.
\newblock \emph{Nature Communications}, 9\penalty0 (1), 2018.

\bibitem[{\"A}ij{\"o} et~al.(2014){\"A}ij{\"o}, Butty, Chen, Salo, Tripathi,
  Burge, Lahesmaa, and L{\"a}hdesm{\"a}ki]{aijo2014methods}
T.~{\"A}ij{\"o}, V.~Butty, Z.~Chen, V.~Salo, S.~Tripathi, C.~B. Burge,
  R.~Lahesmaa, and H.~L{\"a}hdesm{\"a}ki.
\newblock Methods for time series analysis of {RNA}-seq data with application
  to human {Th17} cell differentiation.
\newblock \emph{Bioinformatics}, 30\penalty0 (12), 2014.

\bibitem[Bair et~al.(2006)Bair, Hastie, Paul, and
  Tibshirani]{bair2006prediction}
E.~Bair, T.~Hastie, D.~Paul, and R.~Tibshirani.
\newblock Prediction by supervised principal components.
\newblock \emph{Journal of the American Statistical Association (JASA)},
  101\penalty0 (473):\penalty0 119--137, 2006.

\bibitem[Belsky et~al.(2015)Belsky, Caspi, Houts, Cohen, Corcoran, Danese,
  Harrington, Israel, Levine, Schaefer, et~al.]{belsky2015quantification}
D.~W. Belsky, A.~Caspi, R.~Houts, H.~J. Cohen, D.~L. Corcoran, A.~Danese,
  H.~Harrington, S.~Israel, M.~E. Levine, J.~D. Schaefer, et~al.
\newblock Quantification of biological aging in young adults.
\newblock \emph{Proceedings of the National Academy of Sciences}, 112\penalty0
  (30), 2015.

\bibitem[Bendall et~al.(2014)Bendall, Davis, Amir, Tadmor, Simonds, Chen,
  Shenfeld, Nolan, and Pe'er]{bendall2014single}
S.~C. Bendall, K.~L. Davis, E.~D. Amir, M.~D. Tadmor, E.~F. Simonds, T.~J.
  Chen, D.~K. Shenfeld, G.~P. Nolan, and D.~Pe'er.
\newblock Single-cell trajectory detection uncovers progression and regulatory
  coordination in human {B} cell development.
\newblock \emph{Cell}, 157\penalty0 (3):\penalty0 714--725, 2014.

\bibitem[Bernstein and Sheldon(2016)]{bernstein2016consistently}
G.~Bernstein and D.~Sheldon.
\newblock Consistently estimating {M}arkov chains with noisy aggregate data.
\newblock In \emph{Artificial Intelligence and Statistics (AISTATS)}, pages
  1142--1150, 2016.

\bibitem[Borkan and Norris(1980)]{borkan1980assessment}
G.~A. Borkan and A.~H. Norris.
\newblock Assessment of biological age using a profile of physical parameters.
\newblock \emph{Journal of Gerontology}, 35\penalty0 (2):\penalty0 177--184,
  1980.

\bibitem[Campbell and Yau(2017)]{campbell2017uncovering}
K.~Campbell and C.~Yau.
\newblock Uncovering genomic trajectories with heterogeneous genetic and
  environmental backgrounds across single-cells and populations.
\newblock \emph{bioRxiv}, 2017.

\bibitem[Chen et~al.(2016)Chen, Marioni, Colicino, Peters, Ward-Caviness, Tsai,
  Roetker, Just, Demerath, Guan, et~al.]{chen2016dna}
B.~H. Chen, R.~E. Marioni, E.~Colicino, M.~J. Peters, C.~K. Ward-Caviness,
  P.~Tsai, N.~S. Roetker, A.~C. Just, E.~W. Demerath, W.~Guan, et~al.
\newblock {DNA} methylation-based measures of biological age: meta-analysis
  predicting time to death.
\newblock \emph{Aging (Albany NY)}, 8\penalty0 (9), 2016.

\bibitem[Collado(1997)]{collado1997estimating}
M.~D. Collado.
\newblock Estimating dynamic models from time series of independent
  cross-sections.
\newblock \emph{Journal of Econometrics}, 82\penalty0 (1):\penalty0 37--62,
  1997.

\bibitem[Comfort(1969)]{comfort1969test}
A.~Comfort.
\newblock Test-battery to measure ageing-rate in man.
\newblock \emph{The Lancet}, 294\penalty0 (7635):\penalty0 1411--1415, 1969.

\bibitem[Daniels and Velikova(2010)]{daniels2010monotone}
H.~Daniels and M.~Velikova.
\newblock Monotone and partially monotone neural networks.
\newblock \emph{IEEE Transactions on Neural Networks}, 21\penalty0
  (6):\penalty0 906--917, 2010.

\bibitem[Di et~al.(2009)Di, Crainiceanu, Caffo, and Punjabi]{di2009multilevel}
C.~Di, C.~M. Crainiceanu, B.~S. Caffo, and N.~M. Punjabi.
\newblock Multilevel functional principal component analysis.
\newblock \emph{The Annals of Applied Statistics}, 3\penalty0 (1), 2009.

\bibitem[EuYu(2012)]{euyu2012monomial}
EuYu.
\newblock A non-negative matrix has a non-negative inverse. {What} other
  properties does it have?
\newblock \url{https://math.stackexchange.com/q/214401}, 2012.

\bibitem[Fry et~al.(2017)Fry, Littlejohns, Sudlow, Doherty, Adamska, Sprosen,
  Collins, and Allen]{fry2017comparison}
A.~Fry, T.~J. Littlejohns, C.~Sudlow, N.~Doherty, L.~Adamska, T.~Sprosen,
  R.~Collins, and N.~E. Allen.
\newblock Comparison of sociodemographic and health-related characteristics of
  {UK Biobank} participants with those of the general population.
\newblock \emph{American Journal of Epidemiology}, 186\penalty0 (9):\penalty0
  1026--1034, 2017.

\bibitem[Furukawa et~al.(1975)Furukawa, Inoue, Kajiya, Inada, Takasugi, Fukui,
  Takeda, and Abe]{furukawa1975assessment}
T.~Furukawa, M.~Inoue, F.~Kajiya, H.~Inada, S.~Takasugi, S.~Fukui, H.~Takeda,
  and H.~Abe.
\newblock Assessment of biological age by multiple regression analysis.
\newblock \emph{Journal of Gerontology}, 30\penalty0 (4):\penalty0 422--434,
  1975.

\bibitem[Goodpaster et~al.(2006)Goodpaster, Park, Harris, Kritchevsky, Nevitt,
  Schwartz, Simonsick, Tylavsky, Visser, and Newman]{goodpaster2006loss}
B.~H. Goodpaster, S.~W. Park, T.~B. Harris, S.~B. Kritchevsky, M.~Nevitt, A.~V.
  Schwartz, E.~M. Simonsick, F.~A. Tylavsky, M.~Visser, and A.~B. Newman.
\newblock The loss of skeletal muscle strength, mass, and quality in older
  adults: the health, aging and body composition study.
\newblock \emph{The Journals of Gerontology Series A: Biological Sciences and
  Medical Sciences}, 61\penalty0 (10):\penalty0 1059--1064, 2006.

\bibitem[Greven et~al.(2011)Greven, Crainiceanu, Caffo, and
  Reich]{greven2011longitudinal}
S.~Greven, C.~Crainiceanu, B.~Caffo, and D.~Reich.
\newblock Longitudinal functional principal component analysis.
\newblock \emph{Recent Advances in Functional Data Analysis and Related
  Topics}, 2011.

\bibitem[Gupta and Bar-Joseph(2008)]{gupta2008extracting}
A.~Gupta and Z.~Bar-Joseph.
\newblock Extracting dynamics from static cancer expression data.
\newblock \emph{IEEE/ACM Transactions on Computational Biology and
  Bioinformatics (TCBB)}, 5\penalty0 (2):\penalty0 172--182, 2008.

\bibitem[Gupta et~al.(2016)Gupta, Cotter, Pfeifer, Voevodski, Canini, Mangylov,
  Moczydlowski, and Esbroeck]{gupta2016monotonic}
M.~Gupta, A.~Cotter, J.~Pfeifer, K.~Voevodski, K.~Canini, A.~Mangylov,
  W.~Moczydlowski, and A.~V. Esbroeck.
\newblock Monotonic calibrated interpolated look-up tables.
\newblock \emph{Journal of Machine Learning Research (JMLR)}, 17\penalty0
  (1):\penalty0 3790--3836, 2016.

\bibitem[Hashimoto et~al.(2016)Hashimoto, Gifford, and
  Jaakkola]{hashimoto2016learning}
T.~Hashimoto, D.~Gifford, and T.~Jaakkola.
\newblock Learning population-level diffusions with generative {RNNs}.
\newblock In \emph{International Conference on Machine Learning (ICML)}, pages
  2417--2426, 2016.

\bibitem[Hawkins and Han(2000)]{hawkins2000estimating}
D.~Hawkins and C.~Han.
\newblock Estimating transition probabilities from aggregate samples plus
  partial transition data.
\newblock \emph{Biometrics}, 56\penalty0 (3):\penalty0 848--854, 2000.

\bibitem[Hoffmann et~al.(2015)Hoffmann, Nabbe, and van~den
  Broek]{hoffmann2015effect}
J.~J. Hoffmann, K.~C. Nabbe, and N.~M. van~den Broek.
\newblock Effect of age and gender on reference intervals of red blood cell
  distribution width ({RDW}) and mean red cell volume ({MCV}).
\newblock \emph{Clinical Chemistry and Laboratory Medicine (CCLM)}, 53\penalty0
  (12), 2015.

\bibitem[Horvath(2013)]{horvath2013dna}
S.~Horvath.
\newblock {DNA} methylation age of human tissues and cell types.
\newblock \emph{Genome Biology}, 14\penalty0 (10), 2013.

\bibitem[Jang et~al.(2017)Jang, Gu, and Poole]{jang2017categorical}
E.~Jang, S.~Gu, and B.~Poole.
\newblock Categorical reparameterization with {G}umbel-softmax.
\newblock \emph{arXiv preprint arXiv:1611.01144}, 2017.

\bibitem[Jonsen et~al.(2005)Jonsen, Flemming, and Myers]{jonsen2005robust}
I.~D. Jonsen, J.~M. Flemming, and R.~A. Myers.
\newblock Robust state--space modeling of animal movement data.
\newblock \emph{Ecology}, 86\penalty0 (11):\penalty0 2874--2880, 2005.

\bibitem[Kalbfleisch and Lawless(1984)]{kalbfleisch1984least}
J.~D. Kalbfleisch and J.~F. Lawless.
\newblock Least-squares estimation of transition probabilities from aggregate
  data.
\newblock \emph{Canadian Journal of Statistics}, 12\penalty0 (3):\penalty0
  169--182, 1984.

\bibitem[King(2013)]{king2013solution}
G.~King.
\newblock \emph{A Solution to the Ecological Inference Problem: Reconstructing
  Individual Behavior from Aggregate Data}.
\newblock Princeton University Press, 2013.

\bibitem[Kingma and Ba(2014)]{kingma2014adam}
D.~Kingma and J.~Ba.
\newblock Adam: A method for stochastic optimization.
\newblock \emph{arXiv preprint arXiv:1412.6980}, 2014.

\bibitem[Kingma and Welling(2014)]{kingma2014variational}
D.~P. Kingma and M.~Welling.
\newblock Auto-encoding variational {B}ayes.
\newblock \emph{arXiv preprint arXiv:1312.6114}, 2014.

\bibitem[Klemera and Doubal(2006)]{klemera2006new}
P.~Klemera and S.~Doubal.
\newblock A new approach to the concept and computation of biological age.
\newblock \emph{Mechanisms of Ageing and Development}, 127\penalty0
  (3):\penalty0 240--248, 2006.

\bibitem[Kraemer et~al.(2000)Kraemer, Yesavage, Taylor, and
  Kupfer]{kraemer2000can}
H.~C. Kraemer, J.~A. Yesavage, J.~L. Taylor, and D.~Kupfer.
\newblock How can we learn about developmental processes from cross-sectional
  studies, or can we?
\newblock \emph{American Journal of Psychiatry}, 157\penalty0 (2):\penalty0
  163--171, 2000.

\bibitem[Kumar et~al.(2017)Kumar, Tan, and Cahan]{kumar2017understanding}
P.~Kumar, Y.~Tan, and P.~Cahan.
\newblock Understanding development and stem cells using single cell-based
  analyses of gene expression.
\newblock \emph{Development}, 144\penalty0 (1):\penalty0 17--32, 2017.

\bibitem[Lane et~al.(2016)Lane, Vlasac, Anderson, Kyle, Dixon, Bechtold, Gill,
  Little, Luik, Loudon, et~al.]{lane2016genome}
J.~M. Lane, I.~Vlasac, S.~G. Anderson, S.~D. Kyle, W.~G. Dixon, D.~A. Bechtold,
  S.~Gill, M.~A. Little, A.~Luik, A.~Loudon, et~al.
\newblock Genome-wide association analysis identifies novel loci for chronotype
  in 100,420 individuals from the {UK} biobank.
\newblock \emph{Nature Communications}, 7, 2016.

\bibitem[Levine(2012)]{levine2012modeling}
M.~E. Levine.
\newblock Modeling the rate of senescence: can estimated biological age predict
  mortality more accurately than chronological age?
\newblock \emph{Journals of Gerontology Series A: Biomedical Sciences and
  Medical Sciences}, 68\penalty0 (6):\penalty0 667--674, 2012.

\bibitem[Louis et~al.(1986)Louis, Robins, Dockery, Spiro, and
  Ware]{louis1986explaining}
T.~A. Louis, J.~Robins, D.~W. Dockery, A.~Spiro, and J.~H. Ware.
\newblock Explaining discrepancies between longitudinal and cross-sectional
  models.
\newblock \emph{Journal of Clinical Epidemiology}, 39\penalty0 (10):\penalty0
  831--839, 1986.

\bibitem[MacDonald et~al.(2004)MacDonald, Dixon, Cohen, and
  Hazlitt]{macdonald2004biological}
S.~W. MacDonald, R.~A. Dixon, A.~Cohen, and J.~E. Hazlitt.
\newblock Biological age and 12-year cognitive change in older adults: findings
  from the victoria longitudinal study.
\newblock \emph{Gerontology}, 50\penalty0 (2):\penalty0 64--81, 2004.

\bibitem[Maddison et~al.(2016)Maddison, Mnih, and Teh]{maddison2016concrete}
C.~J. Maddison, A.~Mnih, and Y.~W. Teh.
\newblock The concrete distribution: A continuous relaxation of discrete random
  variables.
\newblock \emph{arXiv preprint arXiv:1611.00712}, 2016.

\bibitem[Magwene et~al.(2003)Magwene, Lizardi, and
  Kim]{magwene2003reconstructing}
P.~M. Magwene, P.~Lizardi, and J.~Kim.
\newblock Reconstructing the temporal ordering of biological samples using
  microarray data.
\newblock \emph{Bioinformatics}, 19\penalty0 (7):\penalty0 842--850, 2003.

\bibitem[McClearn(1997)]{mcclearn1997biogerontologic}
G.~E. McClearn.
\newblock Biogerontologic theories.
\newblock \emph{Experimental Gerontology}, 32\penalty0 (1):\penalty0 3--10,
  1997.

\bibitem[Milovanovic et~al.(2004)Milovanovic, Nilsson, and
  J{\"a}remo]{milovanovic2004relationships}
M.~Milovanovic, E.~Nilsson, and P.~J{\"a}remo.
\newblock Relationships between platelets and inflammatory markers in
  rheumatoid arthritis.
\newblock \emph{Clinica {C}himica {A}cta}, 343\penalty0 (1):\penalty0 237--240,
  2004.

\bibitem[Moffitt(1993)]{moffitt1993identification}
R.~Moffitt.
\newblock Identification and estimation of dynamic models with a time series of
  repeated cross-sections.
\newblock \emph{Journal of Econometrics}, 59\penalty0 (1):\penalty0 99--123,
  1993.

\bibitem[Mozaffarian et~al.(2011)Mozaffarian, Hao, Rimm, Willett, and
  Hu]{mozaffarian2011changes}
D.~Mozaffarian, T.~Hao, E.~B. Rimm, W.~C. Willett, and F.~B. Hu.
\newblock Changes in diet and lifestyle and long-term weight gain in women and
  men.
\newblock \emph{New England Journal of Medicine}, 364\penalty0 (25):\penalty0
  2392--2404, 2011.

\bibitem[Nakamura et~al.(1988)Nakamura, Miyao, and
  Ozeki]{nakamura1988assessment}
E.~Nakamura, K.~Miyao, and T.~Ozeki.
\newblock Assessment of biological age by principal component analysis.
\newblock \emph{Mechanisms of Ageing and Development}, 46\penalty0
  (1):\penalty0 1--18, 1988.

\bibitem[Nowell(1976)]{nowell1976clonal}
P.~C. Nowell.
\newblock The clonal evolution of tumor cell populations.
\newblock \emph{Science}, 194\penalty0 (4260):\penalty0 23--28, 1976.

\bibitem[O'Reilly et~al.(2012)O'Reilly, Hoggart, Pomyen, Calboli, Elliott,
  Jarvelin, and Coin]{o2012multiphen}
P.~F. O'Reilly, C.~J. Hoggart, Y.~Pomyen, F.~C. Calboli, P.~Elliott,
  M.~Jarvelin, and L.~J. Coin.
\newblock Multi{P}hen: joint model of multiple phenotypes can increase
  discovery in {GWAS}.
\newblock \emph{PloS One}, 7\penalty0 (5), 2012.

\bibitem[Plas(1983)]{van1983estimation}
A.~P. V.~D. Plas.
\newblock On the estimation of the parameters of {M}arkov probability models
  using macro data.
\newblock \emph{Annals of Statistics}, 1:\penalty0 78--85, 1983.

\bibitem[Putin et~al.(2016)Putin, Mamoshina, Aliper, Korzinkin, Moskalev,
  Kolosov, Ostrovskiy, Cantor, Vijg, and Zhavoronkov]{putin2016deep}
E.~Putin, P.~Mamoshina, A.~Aliper, M.~Korzinkin, A.~Moskalev, A.~Kolosov,
  A.~Ostrovskiy, C.~Cantor, J.~Vijg, and A.~Zhavoronkov.
\newblock Deep biomarkers of human aging: application of deep neural networks
  to biomarker development.
\newblock \emph{Aging}, 8\penalty0 (5), 2016.

\bibitem[Qiu et~al.(2011)Qiu, Gentles, and Plevritis]{qiu2011discovering}
P.~Qiu, A.~J. Gentles, and S.~K. Plevritis.
\newblock Discovering biological progression underlying microarray samples.
\newblock \emph{PLoS Computational Biology}, 7\penalty0 (4), 2011.

\bibitem[Qu and Hu(2011)]{qu2011generalized}
Y.~Qu and B.~Hu.
\newblock Generalized constraint neural network regression model subject to
  linear priors.
\newblock \emph{IEEE Transactions on Neural Networks}, 22\penalty0
  (12):\penalty0 2447--2459, 2011.

\bibitem[Ram(1986)]{ram1986government}
R.~Ram.
\newblock Government size and economic growth: A new framework and some
  evidence from cross-section and time-series data.
\newblock \emph{The American Economic Review}, 76\penalty0 (1):\penalty0
  191--203, 1986.

\bibitem[Reid and Wernisch(2016)]{reid2016pseudotime}
J.~E. Reid and L.~Wernisch.
\newblock Pseudotime estimation: deconfounding single cell time series.
\newblock \emph{Bioinformatics}, 32\penalty0 (19):\penalty0 2973--2980, 2016.

\bibitem[Relethford et~al.(1978)Relethford, Lees, and Byard]{relethford1978use}
J.~H. Relethford, F.~C. Lees, and P.~J. Byard.
\newblock The use of principal components in the analysis of cross-sectional
  growth data.
\newblock \emph{Human Biology}, pages 461--475, 1978.

\bibitem[Rossol et~al.(2012)Rossol, Kraus, Pierer, Baerwald, and
  Wagner]{rossol2012cd14brightcd16}
M.~Rossol, S.~Kraus, M.~Pierer, C.~Baerwald, and U.~Wagner.
\newblock The {CD}14bright{CD}16+ monocyte subset is expanded in rheumatoid
  arthritis and promotes expansion of the {T}h17 cell population.
\newblock \emph{Arthritis \& Rheumatology}, 64\penalty0 (3):\penalty0 671--677,
  2012.

\bibitem[Stanojevic et~al.(2008)Stanojevic, Wade, Stocks, Hankinson, Coates,
  Pan, Rosenthal, Corey, Lebecque, and Cole]{stanojevic2008reference}
S.~Stanojevic, A.~Wade, J.~Stocks, J.~Hankinson, A.~L. Coates, H.~Pan,
  M.~Rosenthal, M.~Corey, P.~Lebecque, and T.~J. Cole.
\newblock Reference ranges for spirometry across all ages: a new approach.
\newblock \emph{American Journal of Respiratory and Critical Care Medicine},
  177\penalty0 (3):\penalty0 253--260, 2008.

\bibitem[Stein and Shakarchi(2011)]{stein2011fourier}
E.~M. Stein and R.~Shakarchi.
\newblock \emph{Fourier Analysis: an Introduction}, volume~1.
\newblock Princeton University Press, 2011.

\bibitem[Sudlow et~al.(2015)Sudlow, Gallacher, Allen, Beral, Burton, Danesh,
  Downey, Elliott, Green, Landray, et~al.]{sudlow2015uk}
C.~Sudlow, J.~Gallacher, N.~Allen, V.~Beral, P.~Burton, J.~Danesh, P.~Downey,
  P.~Elliott, J.~Green, M.~Landray, et~al.
\newblock {UK Biobank}: an open access resource for identifying the causes of a
  wide range of complex diseases of middle and old age.
\newblock \emph{PLoS Medicine}, 12\penalty0 (3), 2015.

\bibitem[Trapnell et~al.(2014)Trapnell, Cacchiarelli, Grimsby, Pokharel, Li,
  Morse, Lennon, Livak, Mikkelsen, and Rinn]{trapnell2014dynamics}
C.~Trapnell, D.~Cacchiarelli, J.~Grimsby, P.~Pokharel, S.~Li, M.~Morse, N.~J.
  Lennon, K.~J. Livak, T.~S. Mikkelsen, and J.~L. Rinn.
\newblock The dynamics and regulators of cell fate decisions are revealed by
  pseudotemporal ordering of single cells.
\newblock \emph{Nature Biotechnology}, 32\penalty0 (4), 2014.

\bibitem[Waddington(1940)]{waddington1940organisers}
C.~H. Waddington.
\newblock \emph{Organisers and Genes}.
\newblock University Press; Cambridge, 1940.

\bibitem[Wain et~al.(2015)Wain, Shrine, Miller, Jackson, Ntalla, Artigas,
  Billington, Kheirallah, Allen, Cook, et~al.]{wain2015novel}
L.~V. Wain, N.~Shrine, S.~Miller, V.~E. Jackson, I.~Ntalla, M.~S. Artigas,
  C.~K. Billington, A.~K. Kheirallah, R.~Allen, J.~P. Cook, et~al.
\newblock Novel insights into the genetics of smoking behaviour, lung function,
  and chronic obstructive pulmonary disease ({UK} {B}ileve): a genetic
  association study in {UK} {B}iobank.
\newblock \emph{The Lancet Respiratory Medicine}, 3\penalty0 (10):\penalty0
  769--781, 2015.

\bibitem[Wang et~al.(2018)Wang, Dai, Kong, Ma, Erfani, Bailey, Xia, Song, and
  Zha]{wang2018learning}
Y.~Wang, B.~Dai, L.~Kong, X.~Ma, S.~M. Erfani, J.~Bailey, S.~Xia, L.~Song, and
  H.~Zha.
\newblock Learning deep hidden nonlinear dynamics from aggregate data.
\newblock In \emph{Uncertainty in Artificial Intelligence (UAI)}, 2018.

\bibitem[You et~al.(2017)You, Ding, Canini, Pfeifer, and Gupta]{you2017deep}
S.~You, D.~Ding, K.~Canini, J.~Pfeifer, and M.~Gupta.
\newblock Deep lattice networks and partial monotonic functions.
\newblock In \emph{Advances in Neural Information Processing Systems
  (NeurIPS)}, pages 2985--2993, 2017.

\end{thebibliography}

\clearpage
\newpage

\appendix
\section{Proofs}
\label{sec:proofs}

\subsection{Non-negative and monomial matrices}\label{sec:monomial-mat}

In this section, we show that if the inverse of a non-negative matrix $A$ exists and is itself non-negative, then $A$ has to be a monomial matrix.
This is a known linear algebra fact; we provide a proof for completeness,
adapted from \citep{euyu2012monomial}.

\begin{defn}
A matrix $A$ is called a \emph{non-negative} matrix if all of its elements are ${\geq 0}$, and a \emph{positive} matrix if all of its elements are ${>0}$.
\end{defn}

\begin{defn}
A matrix $A$ is called a \emph{monomial} matrix if it has exactly
one non-zero entry in each row and each column.
In other words, it has the same sparsity pattern as a permutation matrix,
though the non-zero elements are allowed to differ from one.
\end{defn}

\newcommand{\iA}{A^{-1}}
\begin{lemma}\label{lem:inv-pos-monomial}
If $A$ is an invertible non-negative matrix and $\iA$ is also non-negative,
then $A$ must be a non-negative monomial matrix.
\end{lemma}
\begin{proof}
Since $A$ is invertible, every row of $A$ must have at least one non-zero element. Consider the $i$-th row of $A$, and pick $j$ such that $A_{ij} \neq 0$. Since $A \iA = I$, we have that the dot product of the $i$-th row of $A$ with the $k$-th column of $\iA$ must be 0 for all $i \neq k$. As $A$ and $\iA$ are both non-negative, this dot product can only be 0 if every term in it is 0, including the product of $A_{ij}$ with $\iA_{jk}$. However, $A_{ij} \neq 0$ by construction, so $\iA_{jk}$ must be 0 for all $i \neq k$. In other words, the $j$-row of $\iA$ must be all 0 except for $\iA_{ji}$.

Applying a symmetric argument, we conclude that the $i$-th row of $A$ must be all 0 except for $A_{ij}$.
Since this holds for all $i$, we have that $A$ must have exactly one non-zero in each row. Moreover, these non-zeros must appear in distinct columns, else $A$ would be singular. We thus conclude that $A$ must be a monomial matrix.
\end{proof}

\subsection{The Jacobians of monotone and order isomorphic functions}\label{sec:propmono}

We recall the definition of monotone and order isomorphic functions from the main text:

\begin{repdefn}{defn:monotone}
A function $f$ is \emph{monotone} if
$u \preceq v \implies f(u) \preceq f(v)$ for all $u, v \in \text{dom}(f)$,
where ordering is taken with respect to the positive orthant
(i.e., $u \preceq v$ means $u_i \leq v_i$ for all $i$).
\end{repdefn}

\begin{repdefn}{defn:order-iso}
An injective function $f$ is an \emph{order isomorphism} if $f$ and $f^{-1}$ restricted to the image of $f$ are both monotone, that is,
$u \preceq v \iff f(u) \preceq f(v)$.
\end{repdefn}

In this section, we establish that monotonicity and order isomorphism impose strong constraints on the function Jacobians.

\begin{lemma}\label{lem:monotone-jacobian}
If a function $f \colon \R^{k_{\text{r}}} \to \R^{k_{\text{r}}}$ is twice differentiable and monotone, then the Jacobian of $f$ evaluated at any $z \in \R^{k_{\text{r}}}$ is a non-negative matrix.
\end{lemma}
\begin{proof}
Assume for contradiction that $f$ is differentiable and monotone,
but that there exists some $z \in \R^{k_{\text{r}}}$ such that the Jacobian
$J_f(z)$ is not a non-negative matrix.
By definition, this implies that we can find $i$ and $j$ such that the $ij$-th entry of $J_f(z)$ is negative.

Let $e_j$ represent the $j$-th unit vector.
By the remainder bound for Taylor approximations, twice differentiability implies that for any compact ball around $z$, we can find some constant $M$ such that
we can write $f(z + \delta e_j) \leq f(z) + \delta J_f(z) e_j + \frac{M}{2}\delta^2 $.
If we pick $\delta < 2|J_f(z)_{ij}|/M$, the first order term dominates.
Since the $ij$-th entry is negative,
this means that $f_i(z + \delta e_j) < f_i(z)$
even though $z + \delta e_j \succeq z$,
contradicting the monotonicity of $f$.
\end{proof}

\begin{lemma}\label{lem:monomial}
If $\diffmap \colon \mathbb{R}^{k_{\text{r}}} \to \R^{k_{\text{r}}}$ is twice continuously differentiable and an order isomorphism, then the Jacobian matrix $J_h(z)$ is a non-negative monomial matrix for all $z \in \R^{k_{\text{r}}}$.
\end{lemma}
\begin{proof}
If $\diffmap$ is an order isomorphism, then $\diffmap$ and $\diffmap^{-1}$ are both monotone.
By \reflem{monotone-jacobian}, their respective Jacobian matrices are non-negative everywhere.

Now, for any $z \in \R^{k_{\text{r}}}$, the inverse function theorem tells us that
${[J_\diffmap(z)]}^{-1} = J_{\diffmap^{-1}}(\diffmap(z))$,
so both $J_\diffmap(z)$ and its inverse ${[J_\diffmap(z)]}^{-1}$ are non-negative.
Applying \reflem{inv-pos-monomial} gives us that $J_\diffmap(z)$ is a non-negative monomial matrix.
\end{proof}

\subsection{Component-wise monotonicity of order isomorphisms}\label{sec:lem-iso-perm-coord}

The conditions on the Jacobian of a twice differentiable order isomorphic function $\diffmap$ imply a constrained form.

\begin{replemma}{lem:iso-perm-coord}[restated]
If $\diffmap\colon \mathbb{R}^{k_{\text{r}}} \to \mathbb{R}^{k_{\text{r}}}$ is an order isomorphism and twice continuously differentiable, $\diffmap$ must be expressible as a permutation followed by a component-wise strictly monotone transform.
\end{replemma}
\begin{proof}
Since $\diffmap$ is bijective, $\diffmap^{-1}$ exists everywhere, which implies that $J_\diffmap(r)$ must have full rank everywhere.
Since $J_\diffmap(r)$ is a monomial matrix by \reflem{monomial}, this means that the sparsity pattern of $J_\diffmap(r)$ cannot vary with $r$;
otherwise, by the intermediate value theorem, there will be some $r$ where $J_\diffmap(r)$ where a row has greater than one nonzero or no nonzeros and thus is not monomial. By definition, a monomial matrix can be decomposed into a positive diagonal matrix and a permutation. Applying the fundamental theorem of calculus to each diagonal entry recovers the strictly monotone tranform, and the permutation matrix defines the permutation. The existence of the antiderivative is guaranteed by construction of $J_\diffmap$ as the derivative of $\diffmap$.
\end{proof}

\subsection{Identifiability in the noiseless setting}\label{sec:prop-identifiability}

We start by establishing two helpful lemmas:

\begin{lemma}\label{lem:composition}
If functions $f_1$ and $f_2$ are both monotone, then $f_1 \circ f_2$ is also monotone.

If $f_1$ and $f_2$ are both bijective order isomorphisms, then $\diffmap \eqdef f_2^{-1} \circ f_1$ is also a bijective order isomorphism.
\end{lemma}
\begin{proof}
The first part of the lemma follows from the transitivity of partial orders: $x \prec y \implies f_1(x) \prec f_1(y) \implies f_2(f_1(x)) \prec f_2(f_1(x))$.

For the second part, note that $\diffmap$ is bijective because it is the composition of two bijective functions.
Now, since $f_1$ and $f_2$ are both order isomorphisms, we know that $f_1, f_1^{-1}, f_2$, and $f_2^{-1}$ are all monotone.
By the first part of the lemma, we conclude that $\diffmap = f_2^{-1} \circ f_1$ and $\diffmap^{-1} = f_1^{-1} \circ f_2$ are both monotone,
implying that $\diffmap$ is an order isomorphism.
\end{proof}

\begin{lemma}
\label{lem:univariate-identity}
If a continuous, univariate, strictly monotone function $\diffmap_i$ is measure preserving for a random variable $x$, $\diffmap_i$ must be the identity map (on the support of $x$).
\end{lemma}
\begin{proof}

  By strict monotonicity, $c_1 < c_2$ implies $\diffmap(c_1) < \diffmap(c_2)$ and thus the CDF is preserved implying that $P(x < c) = P(\diffmap(x) < \diffmap(c)) = P(x < \diffmap(c))$. The last step follows from measure preservation of $\diffmap$.

  Now assume for contradiction that $\diffmap_i$ is not the identity map. We can then pick some $c$ such that $\diffmap(c) \neq c$ and $P(c) > 0$. This implies that $P(x < \diffmap(c)) \neq P(x < c)$ which is a contradiction.
\end{proof}

We can now state and prove identifiability in the noiseless setting:
\begin{repproposition}{prop:identifiability}[restated]
If $f_1$ and $f_2$ and their inverses are twice continuously differentiable and order-isomorphic functions such that $f_1(tr)\disteq f_2(tr)\disteq x_t$ for some $t > 0$, then $f_1$ and $f_2$ are identical up to a permutation.
\end{repproposition}

\begin{proof}
We consider the difference map $\diffmap \eqdef f_2^{-1} \circ f_1$,
which maps latent rates of aging implied by $f_1$ to that of $f_2$.
Our aim is to show that $\diffmap$ must be a permutation, which will give the desired result.

From \reflem{composition}, we know that $\diffmap$ is itself an order isomorphism.
Thus, by \reflem{iso-perm-coord}, it must be expressible as the composition
of a component-wise strictly monotone map and a permutation.

We can further show that this component-wise strictly monotone transformation
has to be the identity transformation.
Since both $f_1$ and $f_2$ map $\ttimesr \mapsto x_t$, $\diffmap$ is measure preserving on $\ttimesr$.
In other words, it maps the probability distribution of $\ttimesr$ to itself.
We can therefore apply \reflem{univariate-identity} to conclude that $\diffmap$ can only be a permutation.

Applying $f_2$ to both sides of $\diffmap = f_2^{-1} \circ f_1$, we get that $f_1$ and $f_2$ have to be permutations of each other, as desired.
\end{proof}

\subsection{Checking order isomorphisms}\label{sec:proofs-checking}

\begin{replemma}{lem:injective-linear}[restated]
  Let $\lineartransformation(x) = Ax$, where $A \in \mathbb{R}^{d \times k}$.
  If we can write $A = P\left[\begin{matrix} B \\ C \end{matrix}\right]$ where $P$ is a permutation matrix, $B$ is a non-negative monomial matrix, and $C$ is a non-negative matrix, then $a$ is an order isomorphism.
\end{replemma}
\begin{proof}
  $\lineartransformation$ is monotone since $A$ is non-negative. To verify that the inverse of $\lineartransformation$ over its image is monotone, let $I_k = [ I ; 0] \in \mathbb{R}^{k\times d}$ be the matrix selecting the first $k$ coordinates.
If $Ax \prec Ay$, every coordinate of $Ax$ is smaller than the corresponding coordinate of $Ay$, so we can jointly permute the rows (i.e., left-multiplying by a permutation matrix) or select a subset of coordinates while preserving ordering.
Thus, $Ax \prec Ay \implies I_kP^{-1}Ax \prec I_kP^{-1}Ay$.
By construction, $I_kP^{-1}A = B$ is a non-negative monomial matrix.
Applying a similar permutation argument, we have that $I_kP^{-1}Ax \prec I_kP^{-1}Ay \implies x \prec y$.
\end{proof}
\begin{repproposition}{prop:iso-checking}[restated]
Let $f \colon \mathbb{R}^k \to \mathbb{R}^d = s_2 \circ \lineartransformation \circ s_1$,
where $s_1 \colon \mathbb{R}^k \to \mathbb{R}^k$ and $s_2 \colon \mathbb{R}^d \to \mathbb{R}^d$ are continuous, component-wise monotone transformations,
and $\lineartransformation \colon \mathbb{R}^k \to \mathbb{R}^d$ is a linear transform.
If $a$ satisfies \reflem{injective-linear},
then $f$ is an order isomorphism.
\end{repproposition}
\begin{proof}
The proof follows from the fact that order preservation is transitive.
$\lineartransformation \circ s_1$ is an order isomorphism onto its image, since $s_1$ is an order isomorphism on the entire $\mathbb{R}^k$ and $\lineartransformation$ is order isomorphic onto its image by Lemma~\ref{lem:injective-linear}. Thus for any $x \prec y \iff \lineartransformation(s_1(x))\prec \lineartransformation(s_1(y))$. Since $s_2$ is an order isomorphism on $\mathbb{R}^d$, we have  $x \prec y \iff \lineartransformation(s_1(x))\prec \lineartransformation(s_1(y)) \iff s_2(\lineartransformation(s_1(x))) \prec s_2(\lineartransformation(s_1(y)))$.
  \end{proof}

\section{UK Biobank dataset and processing}
\label{sec:biobank}

\paragraph{Phenotype filtering.} We selected Biobank phenotypes that were measured for a large proportion of the dataset and that captured diverse and important dimensions of aging and general health. After removing phenotypes which were missing data for many people, redundant (e.g., there are multiple measurements of BMI), or discrete (e.g., categorical responses from a survey question), we were left with 52 phenotypes (Table \ref{tab:phenotypes_list}) across the following categories:
spirometry (a measure of lung function), bone density, body type/anthropometry, cognitive function, vital signs (blood pressure and heart rate), physical activity, hand grip strength, and blood test results. By visual inspection, we categorized the 52 phenotypes into monotone features (45/52) and non-monotone features (7/52) for the cross-sectional model. In the combined longitudinal/cross-sectional model, we modeled an additional 8 features as non-monotone because they increased in the longitudinal data but not in the cross-sectional data, or vice versa.

\paragraph{Sample filtering.} We removed individuals with non-European ancestry, as identified from their genetic principal components, as is commonly done in studies of the UK Biobank to minimize spurious correlations with ancestry particularly in genetic analysis \citep{lane2016genome, wain2015novel}. (The vast majority of individuals in UK Biobank are of European ancestry.) We also removed individuals who were missing data in any of our selected phenotypes.

After filtering, we were left with a train/development set of 213,510 individuals; we report all results on a test set of 53,174 individuals not used in model development or selection. While these samples are cross-sectional (with a measurement at only a single timepoint), we have a single longitudinal followup visit for an additional 8,470 individuals, on which we assess longitudinal progression. UK Biobank data contains two followup visits; we use only longitudinal data from the first followup visit (2-6 years after the initial visit), not the second, because some of the phenotypes we use in model fitting were not measured at the second followup.

\paragraph{Phenotype processing.} We normalized each phenotype to have mean 0 and variance 1. In fitting the model, we first transformed all phenotypes so they were positively correlated with age, by multiplying all phenotypes which were not by negative one, so we could assume that monotone features were monotone increasing. However, all results in the paper are shown with the original phenotype signs.

\paragraph{Diseases, mortality, and risk factors.} We examined associations with 91 diseases which were reported by at least 5,000 individuals in the entire UKBB dataset. Diseases were retrospectively assessed via interview (i.e., subjects developed the disease prior to the measurement of $x_{t_0}$). Second, we examined associations between rates of aging and mortality. In contrast to disease status, mortality was measured after $x_{t_0}$ (all subjects were obviously alive when $x_{t_0}$ was measured); thus, examining associations with mortality serves as an indication that rates of aging predict future outcomes. Finally, we examined 5 binary risk factors: whether the individual currently smokes, if they are a heavy drinker, if they are above the 90th percentile in Townsend deprivation index (a measure of low socioeconomic status), if they have type 2 diabetes, and if they report no days of moderate or vigorous exercise in a typical week.

We examined associations between rates of aging and mortality using a Cox proportional hazards model which controlled for age, sex, and the first five genetic principal components. We report the hazard ratios for a one standard-deviation increase in rate of aging. For the 5 binary risk factors and the 91 diseases, we examined associations using a linear regression model, where the dependent variable was the rate of aging and the independent variable was the risk factor or disease. We controlled for age, sex, and the first five genetic principal components. We filtered for associations which passed a statistical significance threshold of $p=0.05$, with Bonferroni correction for the number of tests performed. \reffig{monotonic_feature_interpretations} reports the diseases/risk factors with the largest positive associations and an effect size of a greater than 1\% increase in the rate of aging; if more than five diseases or risk factors pass this threshold, we report the top five.

\begin{table*}[h!]
\caption{UK Biobank features used in model fitting. * denotes features which are modeled as non-monotone in age when fitting the cross-sectional model. ** denotes additional features which are modeled as non-monotone in age when fitting the model which uses both longitudinal and cross-sectional data. All features which are modeled as non-monotone in the cross-sectional analysis are also modeled as non-monotone in the combined longitudinal/cross-sectional model.}
\label{tab:phenotypes_list}
\centering
\small
\begin{tabular}{|l|}
\hline
\textbf{Feature}   \\ \hline
                                     Spirometry: forced vital capacity \\ \hline
                                      Spirometry: peak expiratory flow \\ \hline
               Spirometry: forced expiratory volume in 1 second (FEV1) \\ \hline
                               Bone density: heel bone mineral density \\ \hline
                   Bone density: heel broadband ultrasound attenuation \\ \hline
                      Bone density: heel quantitative ultrasound index \\ \hline
                                        Body type: body fat percentage \\ \hline
                                            Body type: body mass index \\ \hline
                                          Body type: hip circumference \\ \hline
                Body type: impedance of whole body\textsuperscript{**} \\ \hline
                                             Body type: sitting height \\ \hline
                                            Body type: standing height \\ \hline
                                        Body type: waist circumference \\ \hline
                                   Body type: whole body fat free mass \\ \hline
                                        Body type: whole body fat mass \\ \hline
                                      Body type: whole body water mass \\ \hline
            Cognitive function: duration to first press of snap button \\ \hline
           Cognitive function: mean time to correctly identify matches \\ \hline
              Vital signs: diastolic blood pressure\textsuperscript{*} \\ \hline
                           Vital signs: pulse rate\textsuperscript{**} \\ \hline
                                  Vital signs: systolic blood pressure \\ \hline
           Physical activity: days/week of moderate activity (10+ min) \\ \hline
           Physical activity: days/week of vigorous activity (10+ min) \\ \hline
                         Physical activity: days/week walked (10+ min) \\ \hline
                                 Physical activity: time spent driving \\ \hline
      Physical activity: time spent using computer\textsuperscript{**} \\ \hline
                     Physical activity: time spent watching television \\ \hline
                           Hand grip strength: hand grip strength left \\ \hline
                          Hand grip strength: hand grip strength right \\ \hline
                                           Blood: basophil percentage \\ \hline
                                         Blood: eosinophil percentage \\ \hline
                      Blood: haematocrit percentage\textsuperscript{*} \\ \hline
                   Blood: haemoglobin concentration\textsuperscript{*} \\ \hline
                     Blood: high light scatter reticulocyte percentage \\ \hline
                                 Blood: immature reticulocyte fraction \\ \hline
                       Blood: lymphocyte percentage\textsuperscript{*} \\ \hline
                                   Blood: mean corpuscular haemoglobin \\ \hline
 Blood: mean corpuscular haemoglobin concentration\textsuperscript{**} \\ \hline
                    Blood: mean corpuscular volume\textsuperscript{**} \\ \hline
                             Blood: mean platelet  thrombocyte  volume \\ \hline
                                       Blood: mean reticulocyte volume \\ \hline
                                       Blood: mean sphered cell volume \\ \hline
                        Blood: monocyte percentage\textsuperscript{**} \\ \hline
                      Blood: neutrophil percentage\textsuperscript{*} \\ \hline
                                                 Blood: platelet count \\ \hline
                                                  Blood: platelet crit \\ \hline
                                    Blood: platelet distribution width \\ \hline
          Blood: red blood cell  erythrocyte  count\textsuperscript{*} \\ \hline
                Blood: red blood cell  erythrocyte  distribution width \\ \hline
                         Blood: reticulocyte count\textsuperscript{**} \\ \hline
                    Blood: reticulocyte percentage\textsuperscript{**} \\ \hline
          Blood: white blood cell  leukocyte  count\textsuperscript{*} \\ \hline
\end{tabular}
\end{table*}

\section{Model architecture and hyperparameters}
\label{sec:model-appendix}

\paragraph{Model architecture.} \reffig{model_diagram} illustrates our model architecture. The monotone function $f = s \circ \lineartransformation$
is parametrized as the composition of a monotone elementwise transformation $s \colon \R^{d'} \to \R^{d'}$
with a monotone linear transform $\lineartransformation \colon \R^{k_{\text{r}}} \to \R^{d'}$. We parametrize the linear transformation $\lineartransformation$ using a matrix $A$ constrained to have non-negative entries,
and implement each component $s_i(v) \colon \R_+ \to \R_+$ of $s$
as the sum of positive powers of $v \in \R_+$ with non-negative coefficients
$s_i(v) = \sum_{p_j \in S} w_{j} v^{p_{ij}}$,
where $w_{ij}$ are learned non-negative weights, and $S$ is a hyperparameter.
(For example, $S = [\frac{1}{2}, 1, 2]$ yields the function class $s(v) = w_0 v^{\frac{1}{2}} + w_1 v + w_2 v^2$. We illustrate some of the learned $S$ in Appendix \reffig{fig:nonlinearities}).
We verified that the learned model's $A$ matrix (part of the monotone function $f$)
can be row-permuted into a combination of an approximately monomial matrix and positive matrix,
indicating that we learned an $f$ that was order-isomorphic.

We use neural networks to parametrize the non-monotone functions $\tilde{f}$ and $g$
as well as the encoder (which approximates the posterior over the latent variables $r$ and $b$).
We adopt the following priors:
$r \sim \mathrm{lognormal}(0, \sigma_r^2 I)$;
$b \sim \sN(0, I)$;
and $\epsilon \sim \sN(0, \sigma_\epsilon^2 I)$.
We use a lognormal distribution for $r$ to ensure positivity;
set $\sigma_r = 0.1$ to reflect a realistic distribution of the rates of biological aging~\citep{belsky2015quantification};
and optimize over $\sigma_\epsilon$.
Finally, we simply take $t$ to be an individual's age,
although we could also have optimized over some constant $t_0$ and taken $t = \text{age} - t_0$. 

\paragraph{Hyperparameter selection.} We conducted a random search over the encoder architecture, decoder architecture, learning rate, elementwise nonlinearity, and whether there was an elementwise nonlinearity prior to the linear transformation matrix.
We selected a configuration which performed well (as measured by low reconstruction error/high out-of-sample evidence lower bound (ELBO)) across a range of latent state sizes. Our final architecture uses a learning rate of 0.0005, encoder layer sizes of [50, 20] prior to the latent state, and decoder layer sizes of [20, 50].
Our elementwise nonlinearity is parametrized by $s(y) = \sum_{p_i \in S} w_{i} y^{p_i}$, where  $S = [\frac{1}{5}, \frac{1}{4}, \frac{1}{3}, \frac{1}{2}, 1, 2, 3, 4, 5]$. We found that using an elementwise nonlinearity prior to the linear transformation was not necessary in our dataset, so we only used a nonlinearity after the linearity transformation for interpretability and ease in training. We used Adam for optimization \citep{kingma2014adam} and ReLUs as the nonlinearity.

\section{Baselines}
\label{sec:baselines-appendix}

\paragraph{Linear baselines.} We compare to three linear baselines (PCA, contrastive PCA, and mixed criterion PCA), using the same number of dimensions as in the original model ($k_{\text{r}} + k_{\text{b}} = 15$). We compare to PCA because it is commonly used in biological studies \citep{relethford1978use} and serves as a good baseline for reconstruction performance. (We evaluate PCA reconstruction loss both when PCA is provided age as an input, and when it is not; its reconstruction loss is virtually identical regardless). However, because PCA does not naturally isolate age-related variation, a key goal of our analysis, we also compare to two linear baselines which naturally incorporate age information: contrastive PCA \citep{abid2018exploring} and mixed-criterion PCA \citep{bair2006prediction}. 

Contrastive PCA takes as input a \emph{foreground dataset} and a \emph{background dataset}, and finds a set of latent components which maximize variance in the foreground space while minimizing variance in the background space (trading off between the two objectives using a weighting $\alpha$). The latent components $v$ optimize

\begin{align}
\max_{||v|| =1} v^TC_{\text{foreground}}v - \alpha v^TC_{\text{background}}v
\end{align} 

where $C_{\text{foreground}}$ and $C_{\text{background}}$ are the empirical covariance matrices of the foreground and background datasets, respectively. This corresponds to taking the eigenvectors of the matrix $C_{\text{foreground}} - \alpha C_{\text{background}}$. Because we seek to isolate age-related variation, we use as our foreground dataset the entire dataset of Biobank participants (aged 40-69), and as the background set participants aged 40-49. Contrastive PCA will thus identify components which explain variation in the population as a whole but not within participants of similar ages (40-49). Following the original authors, we experiment with a set of weightings $\alpha$ logarithmically spaced between 0.1 and 1,000. We report results with $\alpha=10$ because this weighting reconstructs the data almost as well as PCA but does not learn identical latent dimensions, indicating that the weighting is having an effect; however, the patterns we report in the main text hold with other $\alpha$ as well. 

Mixed-criterion PCA, like contrastive PCA, uses a two-term objective: the PCA objective (weighted by $1 - \alpha$), and a second term (weighted by $\alpha$) which encourages the learned components to correlate with age:

\begin{align}
\max_{||v|| =1} (1 - \alpha) \var (Xv) + \alpha \cov(Xv, t)
\end{align} 

where $X$ is the matrix of observed features and $t$ is age. When $\alpha = 0$, mixed-criterion PCA reduces to standard PCA; when $\alpha = 1$, it learns a single component which is the linear combination of observed features which correlates most strongly with age. We experiment with a range of $\alpha$ and report results with $\alpha=0.99$, because this yields several top principal components which correlate with age; using a significantly smaller $\alpha$ produces results very similar to PCA, and using a significantly larger $\alpha$ produces only a single meaningful component which is strongly correlated with age, severely harming reconstruction performance. 

\paragraph{Non-linear baseline: non-monotone model.} We use the same hyperparameter settings as for the monotone model but remove the constraint that the age decoder must be linear. Thus, all observed features are represented as an arbitrary function of the age latent state $rt$ plus an arbitrary function of the bias latent state $b$. 

\section{Learning from both cross-sectional and longitudinal data}
\label{sec:longitudinal_extension_experiments}

Our model can naturally incorporate
any available longitudinal data by optimizing the 
joint likelihood of the cross-sectional and longitudinal data.
As cross-sectional and longitudinal data can
display different biases \citep{fry2017comparison, louis1986explaining, kraemer2000can},
this can produce models that are less affected by the biases in a particular dataset.

We handle longitudinal data similarly to cross-sectional data,
but with an additional term in the model objective
that captures the expected log-likelihood of observing the longitudinal follow-up $x_{t_1}$
given our posterior of $r$ and $b$. 
We control the relative weighting between cross-sectional and longitudinal data 
with a single parameter $\llon$; when $\llon = 1$, the longitudinal and cross-sectional losses per sample are equally weighted; when $\llon = 0$, the model tries to fit only the cross-sectional data, and when $\llon \gg 1$, the model tries to fit only the longitudinal data.
We fit the longitudinal model using the same model architecture and hyperparameters as the cross-sectional experiments (Appendix \ref{sec:model-appendix}), varying only the longitudinal loss weighting $\llon$.  The loss for cross-sectional samples is the negative evidence lower bound (ELBO), as before. The loss for longitudinal samples has an additional term that captures the expected log-likelihood of observing the longitudinal follow-up $x_{t_1}$ given our posterior of $r$ and $b$. We use the same model architecture as for the cross-sectional model.
In particular, to avoid overfitting on the small number of longitudinal samples, we share the same encoder;
this means that the approximate posterior over $r$ and $b$ for a longitudinal sample is calculated only
using $x_{t_0}$.
Because we have far more cross-sectional samples than longitudinal samples, we train the model by sampling longitudinal batches with replacement, with one longitudinal batch for every cross-sectional batch. In addition to the 7 non-monotonic features used in the cross-sectional experiments, we add an additional 8 features to the non-monotonic list because they increase in longitudinal data and not in cross-sectional data, or vice versa (Table \ref{tab:phenotypes_list}). 

We search over a range of values of $\llon$ and find that test longitudinal loss (i.e., the negative evidence lower bound on the likelihood of $x_{t_0}$ and $x_{t_1}$) is minimized when $\llon = 1$. This indicates that the model achieves the best longitudinal generalization performance by using cross-sectional data and the small amount of available longitudinal data. 
With higher $\llon$, the model overfits to the small longitudinal dataset.
Repeating our longitudinal extrapolation task (\refsec{model_accuracy})
on a held-out test set of 1687 participants with longitudinal data and comparing to the same three benchmarks,
we found that the model with $\llon = 1$ outperforms 
just predicting $x_{t_0}$ on 83\% of people with followups $> 5$ years
(compared to 66\% with purely cross-sectional data, as in \refsec{model_accuracy});
pure reconstruction on 79\% (vs 61\%);
and the average-cross-sectional-change baseline on 80\% (vs. 60\%). 
The longitudinal model also outperforms benchmarks on the full longitudinal dataset (as opposed to just individuals with followups $> 5$ years) by similarly large margins. 
These results illustrate the benefits of methods which exploit both cross-sectional
and longitudinal data. 

\section{Model stability}
\label{sec:robustness_checks}

We evaluated the stability of the learned rates of aging in response to various model and data perturbations. To compare the rates of aging learned by two different models, we defined $\corrbetweenmodels$ as the correlation between the $r^\i$'s learned by the 2 models, averaged over each component of $r$, and maximized over permutations of components We found that learned rates of aging were stable over random seeds and changes to:
\vspace{-1mm}
\begin{enumerate}
\item The number of non-monotone features.
$\corrbetweenmodels$ with the original model remained high even as we tripled the number of non-monotone features from the original 7, to 25 (for which $\corrbetweenmodels = 0.84$). (We did this by removing monotone constraints on randomly chosen features.)

\item Random subsets of training data. Models trained on different subsets, each containing 70\% of the overall data, learned similar rates $r$ (average $\corrbetweenmodels$ of 0.82 between models).

\item The dimensions $k_r$ and $k_b$ of the time-dependent and bias latent variables. When we altered $k_r$, the model learned many of the same rates of aging: e.g., for $k_r = 4$, $\corrbetweenmodels$ with the original model ($k_r = 5$) was 0.89, and for $k_r = 6$ it was 0.92. Results were also stable when we altered $k_b$ and compared to the original $k_b=10$: $\corrbetweenmodels > 0.8$ for $8 \leq k_b \leq 12$.
\end{enumerate}

\onecolumn
\clearpage
\section{Supplementary Figures}\label{sec:supplementary_figures}

\FloatBarrier

\counterwithin{figure}{section}

\FigStar{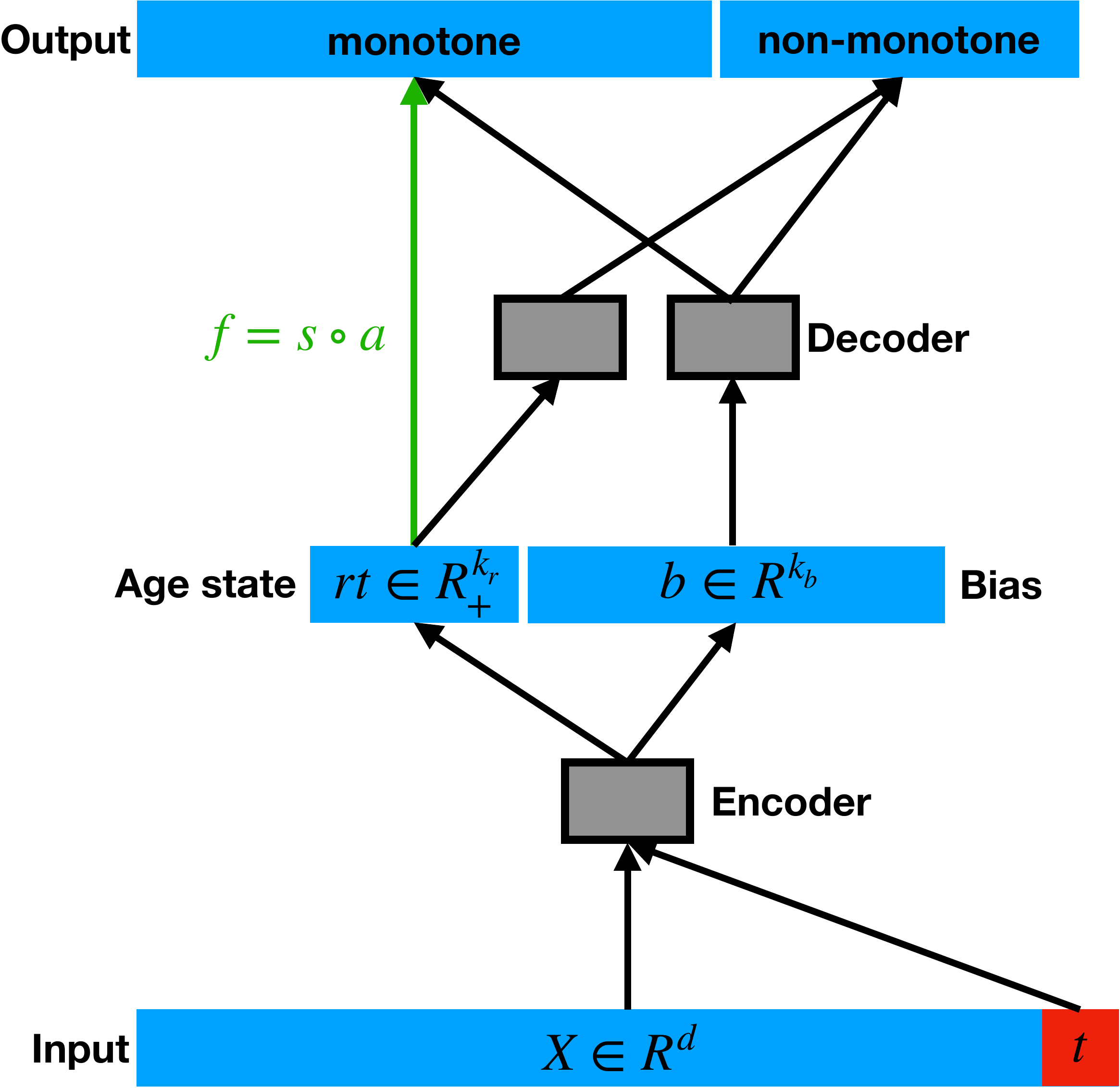}{0.5}{model_diagram}{The model structure. The features $X$ and age $t$ are fed into the encoder to approximate the posterior over the rates of aging $r$ and bias $b$. The grey boxes indicate functions parametrized by neural networks. While both the monotone and non-monotone outputs are a function of both the age state $rt$ and the bias $b$, only the relationship between $rt$ and the monotone outputs (green arrow) is constrained to be monotone and parametrized by $f = s \circ \lineartransformation$.}

\FigStar{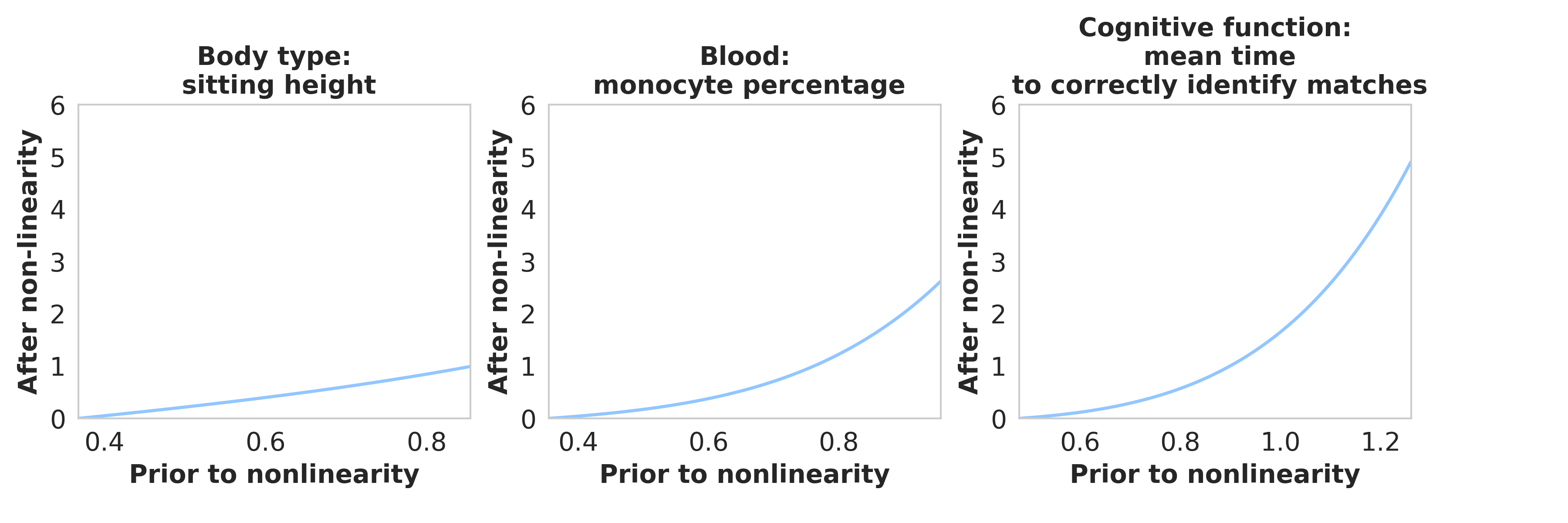}{.1}{fig:nonlinearities}{Representative elementwise transformations $s$. Most elementwise transformations are close to linear, like the left plot, but some are not (right two plots). To determine the relevant domain for each elementwise transformation, we sample latent $rt$ from the fitted cross-sectional model (for $t$ = 40-69), feed it through the linear transformation $a$, and compute the 0.1th and 99.9th percentiles of the resulting distribution for each monotonic feature. This yields the relevant domain over which each elementwise transformation operates.}

\FigStar{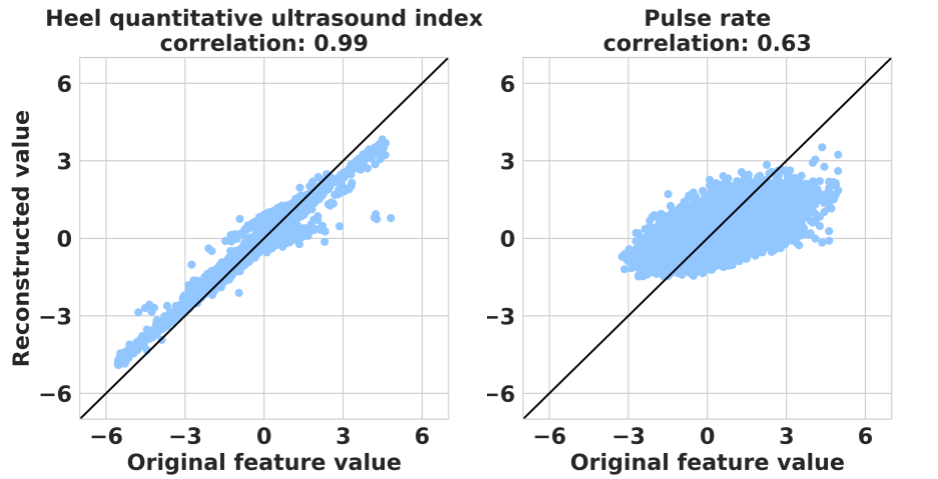}{0.4}{fig:scatter_plots}{Reconstructed vs. actual features. The figure plots the reconstructed $f(rt) + g(b)$ against the actual $x_t$ for the 2 features with the highest ($\rho=0.99$, left) and lowest correlation ($\rho=0.63$, right). Overall, the model fits the data well: reconstructed features are highly correlated with actual features (mean $\rho=0.88$), with most resembling the left plot.}

\FigStar{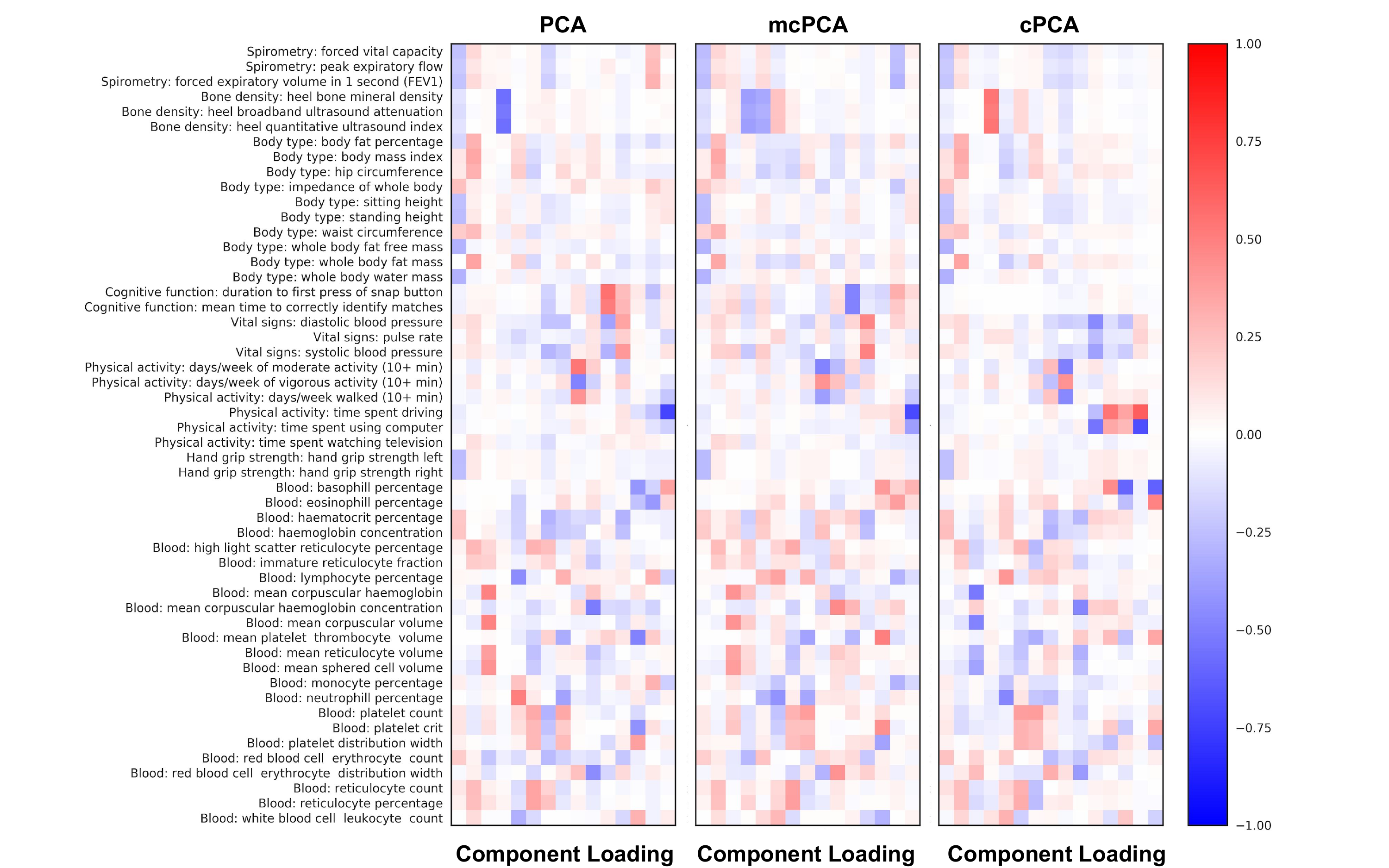}{0.6}{fig:linear_baseline_loadings}{Loadings for the three linear baselines (with 15 latent dimensions) reveal non-sparse latent dimensions which are difficult to interpret and do not clearly differentiate between age and non-age variation. Each cell shows the loading for one component (horizontal axis) and observed feature (vertical axis).}

\twocolumn

\end{document}